\documentclass[acmtog,nonacm]{acmart}

\setcopyright{none}
\renewcommand\footnotetextcopyrightpermission[1]{}

\usepackage{graphicx}
\usepackage{xspace}
\newcommand{\model}{\texorpdfstring{\textbf{SceneRuler}}{SceneRuler}\xspace}

\definecolor{qpYellow}{HTML}{E5B455}
\definecolor{qpOrange}{HTML}{E09A6E}
\definecolor{qpRed}{HTML}{C0564F}
\definecolor{qpPink}{HTML}{DE8C9D}
\definecolor{qpMauve}{HTML}{C58BC0}
\definecolor{qpPurple}{HTML}{9B93C9}
\definecolor{qpLightBlue}{HTML}{7FA6CF}
\definecolor{qpBlue}{HTML}{5B8DB8}
\definecolor{qpTeal}{HTML}{6FB5AE}
\definecolor{qpGreen}{HTML}{92B573}
\definecolor{qpOlive}{HTML}{C2BB66}
\definecolor{qpGrey}{HTML}{B9BEC4}
\definecolor{red}{HTML}{C0564F}
\definecolor{ACMPurple}{HTML}{9B93C9}
\definecolor{ACMDarkBlue}{HTML}{5B8DB8}
\definecolor{ACMBlue}{HTML}{5B8DB8}
\definecolor{ACMLightBlue}{HTML}{7FA6CF}
\definecolor{ACMOrange}{HTML}{E09A6E}
\definecolor{ACMYellow}{HTML}{E5B455}
\definecolor{ACMRed}{HTML}{C0564F}
\definecolor{ACMGreen}{HTML}{92B573}
\colorlet{bestbg}{qpBlue!58!white}
\colorlet{secondbg}{qpBlue!25!white}
\newcommand{\best}[1]{{\setlength{\fboxsep}{1pt}\colorbox{bestbg}{#1}}}
\newcommand{\second}[1]{{\setlength{\fboxsep}{1pt}\colorbox{secondbg}{#1}}}

\begin{document}

\title[\model]{\model: Measuring Physical Quantities in Video via Minimal Sufficient Reconstruction}

\author{Yumeng He}
\affiliation{%
  \institution{University of California, Los Angeles}
  \country{USA}}

\author{Gezheng Kang}
\affiliation{%
  \institution{University of California, Los Angeles}
  \country{USA}}

\author{Fanlian Zeng}
\affiliation{%
  \institution{Envora Inc.}
  \country{USA}}

\author{Yin Yang}
\affiliation{%
  \institution{University of Utah}
  \country{USA}}

\author{Chenfanfu Jiang}
\affiliation{%
  \institution{University of California, Los Angeles}
  \country{USA}}

\begin{abstract}
Many questions about videos ask for a number, such as how tall a person is, how far a cart travels, or how fast a ball leaves a ramp. We study this task, quantitative physical reasoning from video, where the inputs are a video, a question, and a prior of known facts such as object sizes, speeds, and distances from the camera to named objects. End-to-end video-language models lack an explicit metric scale and physical consistency, and their errors cannot be inspected. We present \model, which instead recovers a measurable physical scene and answers with that scene as evidence. Its central principle, Minimal Sufficient Reconstruction, builds, for each question, only the smallest scene that suffices to measure the asked quantity, and tests sufficiency on measurements rather than assuming it. The scene is written as a program whose structure a vision-language model writes and whose magnitudes are left to measurement. Physics-guided Fitting measures them through physical motion laws, so that gravity itself can anchor the metric scale, and marks every parameter, and the read-out of the asked quantity, as determined or undetermined. Each question receives a read-out and an answer. The read-out is the asked quantity evaluated on the scene, and it is empty when the scene does not fix that quantity. The answer is always given by a model that takes the scene as evidence. We evaluate on QuantiPhy and also apply the method to 27 in-the-wild videos that we collected, which have no ground-truth answers.
On its validation split, the final answer reaches a mean relative accuracy of 0.709 against 0.624 for the strongest baseline, and the reconstruction determines the asked quantity on 47 of 158 questions, where the final answer reaches 0.781.
Because every intermediate quantity is explicit, each answer can be traced to the objects, scale, and motion behind it.
\end{abstract}

\keywords{quantitative physical reasoning, video question answering, metric scene reconstruction, Physics-guided Fitting, Minimal Sufficient Reconstruction}

\begin{teaserfigure}
  \centering
  \includegraphics[width=\linewidth]{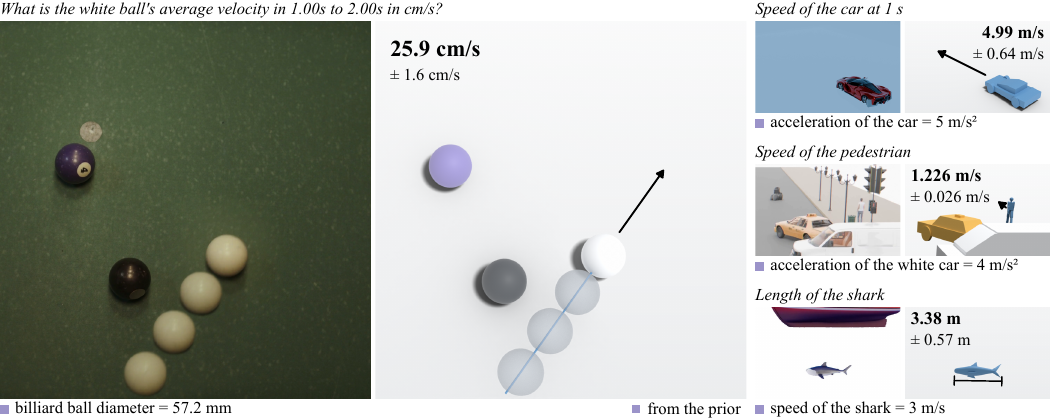}
  \caption{\textbf{\model.} Given a video, a question, and a prior of known facts, \model reconstructs a measurable physical scene and answers with it as evidence. Left, the white ball at the asked instants in the video and the fitted scene rendered from the fitted camera. The prior gives only the ball diameter. The scene carries it to a metric scale, and the velocity is read from the program with its uncertainty. Right, three more scenes, each with its prior.}
  \Description{Left, a video frame of a billiard table with the white ball pasted at four earlier instants, and beside it a rendering of the fitted scene in the same framing, with the white ball, its faint copies, its fitted path in blue, an arrow for its velocity, and the read-out 25.9 plus or minus 1.6 centimetres per second. Right, three smaller pairs of a frame and a rendering, the speed of a red car at one second, the speed of a walking pedestrian at a crossing, and the length of a shark swimming under a boat, each with its read-out and the prior statement under it.}
  \label{fig:teaser}
\end{teaserfigure}

\maketitle

\section{Introduction}
\label{sec:introduction}

How tall is the person on the left? How fast is the ball when it leaves the
ramp? How far does the cart travel in two seconds? Questions like these ask for
a number in meters, meters per second, or meters per second squared, and the
only evidence is an ordinary video together with a few facts the asker already
knows. These facts are the size of one object, the speed of another, or the distance from the
camera to an object at a given time. We study this setting, \emph{quantitative physical
reasoning from video}, and present \emph{\model}, which treats it as a
reconstruction problem. It recovers a \emph{measurable physical scene} from the
video and reads the asked quantity off that scene. The input is always the same three things, a video, a question, and
a prior. The prior collects every known fact that comes with the question.
These facts are the size, speed, or acceleration of a named object, and, where given, the
distance from the camera to a named object.

The direct alternative is to ask a large video-language model for the number, for example Gemini-4-Carbon~\cite{gemini4carbon} or Qwen3.5~\cite{qwen35}.
Such models describe what happens in a video well, but a metric answer needs two
things they do not represent explicitly. The first is a metric scale that ties pixels to
meters. The second is consistency with physics, so that positions, velocities, and
accelerations over time agree with one another and with gravity~\cite{physbench2024,quantiphy2025}.

When an answer is produced end to end, there is also nothing to inspect. One
cannot tell whether an error comes from the wrong object, the wrong scale, or
the wrong motion. A reconstructed scene, by contrast, makes every intermediate
quantity explicit and checkable against the video.

Recent monocular depth estimation~\cite{depthpro2025} and dynamic scene
reconstruction~\cite{monst3r2025,megasam2025} recover dense geometry and camera
motion from casual video. Full metric 3D reconstruction of an arbitrary video is, however, more than a question requires, and the images of a single camera do not fix its metric scale.
Our central idea is \emph{Minimal Sufficient Reconstruction}. For each question,
it reconstructs the smallest scene that suffices to measure the asked quantity, and
nothing more. That scene holds the objects the question and the prior refer to,
what they rest on or hang from, one camera, and their motion, with every object a
shape as coarse as the question allows. Appearance, the background, and objects
that play no role in the answer are left out. Sufficiency is tested on
measurements rather than assumed. A scene is \emph{sufficient} when the observations and the prior determine
the asked quantity, so that nothing it is read from is left open and its own uncertainty is small, and the prior anchors the metric
scale that the images leave open. It stays \emph{minimal} because an element is added only when the question, the
prior, or an element already present needs it, and a parameter the data do not
constrain is reported as undetermined rather than filled in.

To make sufficiency testable, the scene must be explicit down to every number.
\model therefore writes it as a \emph{scene program}, a renderable description
of the objects, their supports, their motion, and the camera, with every prior
fact bound to the quantity it states. The program divides the labor between
models and measurement. A vision-language model (VLM) writes the structure. It decides which
objects exist, what supports what, and which physical law governs each stretch of
motion. Metric values enter only from the prior, a convention such as gravity on
Earth, a commonsense size held softly, or the fit. Models decide what the scene
is. Measurements decide how big and how fast it is.

\emph{Physics-guided Fitting} then measures the magnitudes against the object
masks of all frames at once. Motion is fitted through physical laws rather than
frame by frame, so a few physical parameters absorb many noisy observations, and
these parameters are the quantities the questions ask about. Because free flight
and swinging depend on gravity, gravity becomes an anchor of the metric scale
beside the prior. A fit returns a value for every parameter, so it also states
which values it can vouch for. Every parameter, the metric scale, and the asked
quantity are marked \emph{determined} or \emph{undetermined}.

A structure written by a model can be wrong, for example by a missed stop, a wrong support, or a
wrong depth relation. The fit exposes such errors as numbers, residuals that
drift, stated facts that do not hold, and a read-out that is empty. These
numbers stay with the scene and reach the answer as evidence.

Each question receives a read-out and an answer. The \emph{read-out} is the
asked quantity evaluated on the fitted program. It is empty when the scene does
not determine that quantity. The \emph{final answer} is always given. A VLM sees
the question, the prior, frames of the video, the rendered program, and the
reconstruction as evidence, without being told whether the scene determines the
asked quantity, and returns one number. The read-out thus states, separately from
the answer, whether the reconstruction measures what was asked. Each question is a self-contained
unit. Its scene is built only from the video, that question, and its prior,
without borrowing information from other questions on the same video.

We evaluate on QuantiPhy~\cite{quantiphy2025}, a benchmark of quantitative physical
questions over videos. We also apply the method to 27 in-the-wild videos that we collected, with 54 questions and no ground-truth answers (Section~\ref{sec:exp-qualitative} and Appendix~\ref{app:wild}).
In summary, we contribute:
\begin{itemize}
  \item \emph{Minimal Sufficient Reconstruction}, a principle for quantitative
    video question answering. It reconstructs, per question, the smallest scene
    that measures the asked quantity, with sufficiency tested on measurements
    (Section~\ref{sec:method-msr});
  \item the \emph{scene program}, an explicit scene whose structure a VLM writes
    and whose metric magnitudes enter only from the prior, conventions, class
    sizes, or the fit (Section~\ref{sec:method-program});
  \item \emph{Physics-guided Fitting}, which fits motion through physical laws,
    with gravity as an anchor of the metric scale, and marks every parameter and
    the read-out as determined or undetermined
    (Section~\ref{sec:method-fit});
  \item a read-out and an answer per question. The read-out is empty when the
    scene does not determine the asked quantity, and the answer is always given
    and reads the reconstruction as evidence
    (Section~\ref{sec:method-answer}).
\end{itemize}

\section{Related Work}

\label{sec:related}

\begin{figure*}[t]
  \centering
  \includegraphics[width=\linewidth]{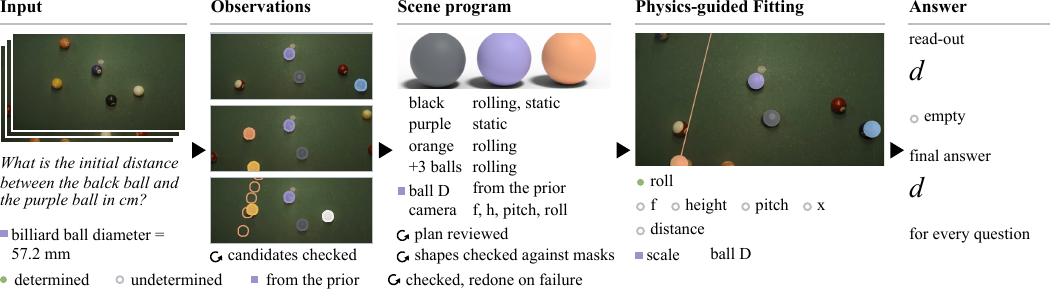}
  \caption{\textbf{Overview}, on one question. The question asks for the initial distance between two billiard balls, with the balls' diameter as the prior. Tracks give one mask per object and frame. The VLM writes the structure of the scene program (Section~\ref{sec:method-program}). The structure holds the objects with coarse solid shapes, their motion as segments with laws, the prior bound to the balls' size, and the camera. Physics-guided Fitting (Section~\ref{sec:method-fit}) measures the magnitudes against the masks and marks each parameter determined (filled) or undetermined (hollow). The candidate tracks, the plan, and the authored shapes are each checked where they are made and redone when the check fails (return arrows). Here the focal length, the camera's height and pitch, and the black ball's position along the table are left open, and so is the asked distance. The read-out is therefore empty. The final answer is given for every question (Section~\ref{sec:method-answer}). Figure~\ref{fig:pipeline-full} in the appendix shows every stage of this example with its numbers.}
  \Description{The figure has five panels from left to right. They show input frames with the question and prior, the same frames with every tracked ball's mask filled in its own colour and the rolling ball's earlier outlines on the last frame, three rendered ball shapes in the same colours and the program structure, small return arrows where the tracks, the plan, and the shapes are checked, the fitted scene rendered onto the video with filled and hollow markers for determined and undetermined parameters, and the scene and final answers.}
  \label{fig:pipeline}
\end{figure*}

\subsection{Quantitative Physical Reasoning}
Vision-language models (VLMs) already carry a good deal of commonsense physical knowledge. From a single image they can estimate how large, how tall, or how heavy an object is, and models trained on metric spatial question answering, such as SpatialVLM~\cite{spatialvlm2024} and SpatialRGPT~\cite{spatialrgpt2024}, estimate distances and sizes in real units. Video adds quantities that no single frame contains, such as velocity and acceleration, and with them the question of whether an estimate is consistent with physical law over time. A line of benchmarks measures how well models do this. CLEVRER~\cite{clevrer2020} and Physion~\cite{physion2021} test qualitative predictions such as collision outcomes and stability. Physion++~\cite{physionplus2023} and PhysBench~\cite{physbench2024} extend the evaluation to latent physical properties such as mass, friction, and elasticity. QuantiPhy~\cite{quantiphy2025} asks for numerical answers, an object's size, velocity, or acceleration, given textual priors about the scene. These works share our goal of reading metric physical quantities from visual input. End-to-end VLMs, however, have no explicit mechanism that ties pixels to metric units or enforces kinematic consistency, so their numerical answers can drift and cannot be inspected. Our method instead measures the asked quantity in an explicitly reconstructed scene whose motions follow physical laws, and reports whether that scene determines the asked quantity.

\subsection{Video-based 3D/4D Scene Reconstruction}
Monocular depth estimation has progressed from high-quality relative depth, as in DPT~\cite{dpt2021} and Depth Anything~\cite{depthanything2024}, to metric depth, as in ZoeDepth~\cite{zoedepth2023} and Depth Pro~\cite{depthpro2025}, which also estimates the focal length. DUSt3R~\cite{dust3r2024} regresses 3D point maps and camera poses from uncalibrated images, and MonST3R~\cite{monst3r2025} and MegaSaM~\cite{megasam2025} extend such reconstruction to dynamic videos, recovering per-frame geometry and camera trajectories. Dense point trackers such as CoTracker~\cite{cotracker2023} supply long-range 2D correspondences. Closest to our setting, PhysMind~\cite{physmind} answers quantitative questions by reconstructing posed metric meshes of the question objects, combining video segmentation, metric depth and intrinsics estimation (MoGe-2~\cite{moge2}), image-to-3D object reconstruction (SAM 3D Objects~\cite{sam3dobjects}), and 6D pose tracking (FoundationPose~\cite{foundationpose2024}), and measures the asked quantity on the meshes. Full 4D reconstruction of a dynamic monocular video is costly and often ill-posed, and errors in parts of the scene that the question never asks about still propagate into the answer. Ours follows the principle of \emph{Minimal Sufficient Reconstruction}. It reconstructs only the objects, supports, camera, metric scale, and motions that the asked quantity depends on, each object as a coarse constructive shape rather than a detailed mesh, and leaves the rest of the scene unmodeled.

\subsection{Agentic Workflows for Modeling}
Tool-using VLM agents decompose visual reasoning into calls to specialized modules. For example, the programs that VisProg~\cite{visprog2023} and ViperGPT~\cite{vipergpt2023} generate invoke detection, segmentation, and depth models and compose their outputs. Closer to our task, Code as Worlds~\cite{codeasworld} has an agent discover executable world representations for physical reasoning, and PhysMind~\cite{physmind} turns a video into an executable world for training-free physical reasoning.
A complementary line builds explicit models of physical dynamics. The Neural Physics Engine~\cite{npe2017} learns compositional object-based dynamics, PhyDNet~\cite{phydnet2020} disentangles PDE-constrained physical dynamics from residual factors for video prediction, and system identification fits parametric motion models, such as projectile flight, rolling, or a pendulum, to observed trajectories~\cite{galileo2015,jaques2020physics,kandukuri2022physical}. Our approach combines the two with a strict division of labor. A VLM writes only the structure of a scene program, while every metric magnitude comes from the prior, a convention, a commonsense size held softly, or the fit, and each object's motion is parameterized by the physical law that governs it instead of being fitted freely frame by frame.

\section{Method}
\label{sec:method}

The input of \model is a video, a question, and a prior. Its output is one
number in the unit the question asks for. The prior is a list of statements. These statements give the
size, speed, or acceleration of a named object, and, where given, the distance
from the camera to a named object at a given time. \model answers by
reconstructing a small measurable scene and answering with the asked quantity
read off it as evidence
(Figure~\ref{fig:pipeline}). Section~\ref{sec:method-msr} states what that scene
must contain. Section~\ref{sec:method-program} writes it as a program whose
structure a vision-language model (VLM) proposes and whose magnitudes are left to
measurement. Section~\ref{sec:method-fit} measures the magnitudes against the
video under the prior and physical law, and decides which of them, and whether
the read-out of the asked quantity, the evidence determines. Section~\ref{sec:method-answer} produces the
answer. Each
question is processed on its own. No other question's text, prior, or answer
enters its scene.

\subsection{Minimal Sufficient Reconstruction}
\label{sec:method-msr}

\begin{figure}[!t]
  \centering
  \includegraphics[width=\columnwidth]{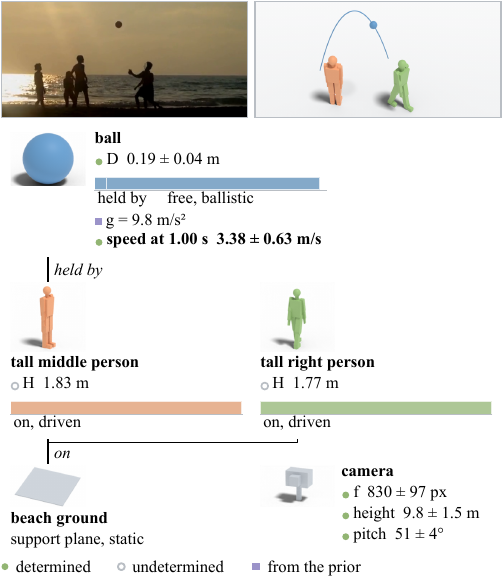}
  \caption{\textbf{One scene program as a tree.} A ball is thrown between two people on a beach, and the question asks for its speed at 1.00\,s with the prior $g = 9.8$\,m/s$^2$. Top, the input frame at 1.00\,s and the fitted program rendered from the fitted camera, each body in the colour of its node. Below, the tree is drawn as the scene stands. The asked ball is on top, held by one person and then free in a ballistic segment that reads $g$, which fixes the scale. The people stand on the ground plane at the base, and the camera stands beside it. Each node shows the fitted shape, its size, and its segments on one time strip.}
  \Description{At the top, a video frame of people on a beach at sunset with a ball in the air, and beside it the fitted scene rendered from the same camera, an orange and a green articulated figure and a blue ball with its arc. Below, a tree read from top to bottom. On top, the ball with a determined diameter, a time strip that is held for a moment and then free and ballistic, the prior g = 9.8 m/s squared, and the determined speed at 1.00 s. An edge labelled held by leads down to the first of two standing people, each with its height and a time strip with one segment. Edges labelled on lead from both people down to the ground plane at the base, with the camera and its focal length, height, and pitch beside it. A key at the bottom reads determined, undetermined, and from the prior.}
  \label{fig:scenetree}
\end{figure}

\begin{figure}[t]
  \centering
  \includegraphics[width=\linewidth]{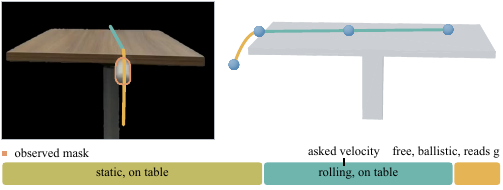}
  \caption{\textbf{Segments and laws.} A ping-pong ball rests on a table, rolls along it, and leaves its edge in ballistic flight. Each segment is a relation and a law with a few parameters. The fitted path is drawn into the video (left) and with the program's bodies (right). The asked velocity is read from the rolling law at the asked time, and the ballistic segment reads $g$, which anchors the scale beside the prior.}
  \Description{The left panel shows a video frame of a table with the ball's fitted path drawn in three colours. The right panel shows the table and the ball at four times as solid shapes with the same coloured path. The bottom shows a time bar split into a static, a rolling, and a ballistic segment, with a mark at the asked time.}
  \label{fig:laws}
\end{figure}

For each question we reconstruct the smallest scene that suffices to measure the
asked quantity. The scene contains the objects that the question and the prior
refer to, the supports and holders they rest on or hang from, one camera, and one
motion law per time segment of each object. Every object is a coarse shape, as
coarse as the question allows. A ball is a sphere and a car a box or a few boxes.
Appearance, the background, and objects that play no role in the answer are left
out.

Both properties are tested on measurements rather than assumed, and this is what
shapes the rest of the method. \emph{Sufficiency} is a property of the fitted
scene. Every parameter is either determined by the observations and the prior or
marked undetermined with a reason, and the asked quantity counts as measured only
when nothing it is read from is left open and its own uncertainty is small
(Section~\ref{sec:method-ident}).
\emph{Minimality} is a property of how the scene is built. An element enters only
when the question, the prior, or an element already present needs it, and the fit
reports a parameter the data do not constrain as undetermined rather than filling
it in. A scene that is too small shows up as residuals and undetermined
parameters. A scene that is too large shows up as parameters the data do not
constrain.

\subsection{The Scene as a Program}
\label{sec:method-program}

To make sufficiency testable, the scene must be explicit down to every number. We
therefore write it as a \emph{scene program}. It consists of named parameters, objects, and one
camera, which can be rendered on every frame and compared with the video.
Figure~\ref{fig:scenetree} draws one program as a tree, the form of the scene
graphs of earlier work~\cite{johnson2015image,armeni20193d,rosinol20203d}.
The support plane is the root, and each body hangs from what it rests on or is held
by. Each body carries its shape and size, its segments with their laws, and the
marks of what the fit determines.

\paragraph{Parameters and their sources.}
Each parameter has a value, a unit, an uncertainty $\sigma$, and a \emph{source}.
The source is \emph{prior} (stated in the prior), \emph{convention} (a standard value taken as
given, such as $g$ on Earth), \emph{class} (a commonsense size, held only softly),
\emph{fit} (constrained by the observations), or \emph{vlm} (a model's estimate,
used only as an initial value). The source rule is the method's central division
of labor. A metric value enters the program only from the prior, a convention, a
class size, or the fit. Models decide what the scene is. Measurements decide how
big and how fast it is.

\paragraph{Objects and motion laws.}
An object is a plane or a body. A body's shape is a coarse constructive solid
built from a few primitives, with its metric extents as scene parameters, so a
size changes without re-authoring the shape. An object's presence over time is a
sequence of segments, each with a relation to an earlier object, such as resting
on a support or being held, and a \emph{motion law}, from static and constant
velocity to ballistic flight, rolling, pendulum, and circular motion
(Appendix~\ref{app:laws}). Physical quantities are thus parameters of laws
(Figure~\ref{fig:laws}). A ballistic segment,
\begin{equation}
  \label{eq:ballistic}
  \mathbf{p}(t)=\mathbf{p}_0+\mathbf{v}_0\,(t-t_0)-\tfrac{1}{2}\,g\,(t-t_0)^2\,\hat{\mathbf{z}},
\end{equation}
has a handful of parameters however many frames it spans, and its initial
velocity is exactly what a question asks about. The camera is one static pinhole
whose focal length and pose above the ground are parameters as well.

\begin{figure}[t]
  \centering
  \includegraphics[width=\columnwidth]{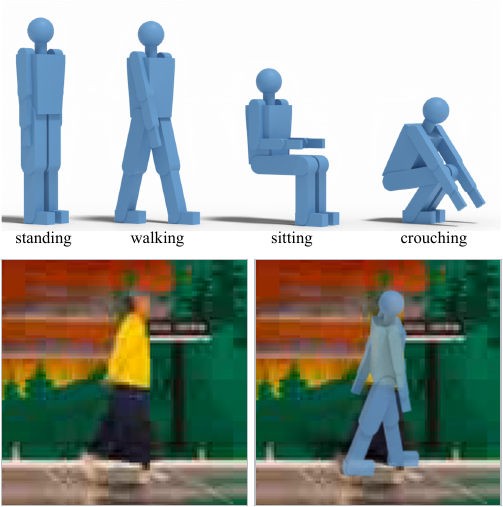}
  \caption{\textbf{The person template.} Top, the four poses at one height. Bottom, a walking person of the benchmark and the template fitted to her.}
  \label{fig:person}
  \Description{Top, four blue articulated figures on one floor, standing, walking, sitting, and crouching, labelled under each. Bottom left, a crop of a video frame with a woman walking in front of a wall. Bottom right, the same crop with the blue walking figure fitted over her.}
\end{figure}

\paragraph{People.}
A person's shape is known in advance up to size and pose, so a human is never
authored by the model. Every person is one articulated template written in code
from its standing height $H$ alone. Every other length is a fixed fraction of $H$
from an anthropometric segment table (Figure~\ref{fig:person}). The plan chooses
only a categorical pose, standing, walking, sitting, or crouching. The fit scales
$H$ and nothing else, so a person adds one scene parameter and no freedom of shape.
Adult stature lies in a narrow range, so $H$ can also carry a commonsense class size and
serve as a soft anchor of the metric scale. Like every parameter, $H$ leaves the
fit marked determined or undetermined, and a person's height is read as $H$ in
every pose (Appendix~\ref{app:laws}).

\paragraph{Observations.}
The program is compared with two fixed inputs. The first is one binary mask per
tracked object and frame, from an open-vocabulary detector~\cite{groundingdino2024}
and a video segmenter~\cite{sam2_2025} (Appendix~\ref{app:inputs}). A candidate
track is kept only if it passes checks of identity, presence, size, and the image
border, and otherwise the next candidate is tried. Each object's
\emph{mask noise} $\sigma_o$, the boundary jitter between consecutive masks, is
the unit in which every pixel threshold of the method is stated. The second input
is the prior, parsed into numbered statements whose values the program never
changes.

\paragraph{The model writes the structure.}
A VLM writes the categorical part of the program from labelled frames, the
question, and the statements. It writes which objects exist and which track each is, their
shapes and proportions, their motion as segments with relations and laws, and
which parameter or derived quantity each prior statement and the question refer
to. Where masks alone leave the scale open, it may add further evidence. This evidence can be bodies at
the same depth, commonsense sizes as soft votes, and image measurements that tie a
statement about an untracked feature, such as a lane's width, to marked points on
the plane it lies on. A second call reviews the plan. It may add objects or
sharpen descriptions but not remove them. All model output is checked in code before it becomes
structure, and every prior statement must be accounted for, so none is dropped
silently (Appendix~\ref{app:plan}). Each authored shape is also checked against
the masks before any fit. Its silhouette is aligned to the observed mask over a
sweep of viewing directions, and the boundary distance that remains is measured.
The VLM judges each body from the aligned outlines and these numbers. A body it
judges wrong is re-authored or given new proportions, and the change is kept only
when the boundary distance falls by more than the mask noise. The number of
rounds per body is capped. The check never reads the question or the fit. The result is a structure whose magnitudes are
free except where the prior or a convention gives them, and which the fit now
measures.

\subsection{Physics-guided Fitting}
\label{sec:method-fit}

Physics-guided Fitting estimates every free parameter of the structure, the
camera, the sizes, and the law parameters, against the masks of all frames at
once. Because motion is fitted through its law rather than frame by frame, a few
physical parameters absorb many noisy observations, and the fitted parameters are
the quantities the questions ask about.

\paragraph{Objective.}
For object $o$ on frame $f$, let $d_{of}(\theta)$ be the silhouette discrepancy
between the rendered program with parameters $\theta$ and the observed mask. It is
symmetric, so a model larger than the silhouette costs as much as a smaller one,
and occlusion and the image border are not counted as error
(Appendix~\ref{app:fit}). The prior's statements on derived quantities, such as a
speed or a distance to the camera, together with structural relations, such as
continuity across segments, form a set of \emph{held rows} $e_h(\theta)$. The fit
solves
\begin{equation}
  \label{eq:objective}
  \begin{split}
  \min_{\theta}\;&\sum_{o}\sum_{f\in\mathcal{F}_o}\frac{d_{of}(\theta)}{\sigma_o^2}
  +\sum_{c}\Big(\frac{x_c(\theta)-\bar{x}_c}{\sigma_c}\Big)^2\\
  &+\lambda\,\lVert\theta-\theta^{0}\rVert^2
  \quad\text{s.t.}\quad e_h(\theta)=0\;\;\forall h,
  \end{split}
\end{equation}
where $c$ runs over class votes $\bar{x}_c\pm\sigma_c$ and the last term is a weak
pull toward the initial values, which keeps a direction the data do not see where
it started. A prior statement on a size or on $g$ simply fixes that parameter.

\paragraph{Statements are targets.}
A stated value is a target, not an interval. The constraint in
Eq.~\eqref{eq:objective} is enforced by projecting the parameters onto the held
rows after every step (Appendix~\ref{app:fit}). Each held
row still has a tolerance, read from the problem rather than chosen as a
constant, such as the written precision of a stated value. The tolerance
describes uncertainty and is used only where uncertainty is the question, namely whether
a row is violated, whether two statements contradict each other, and how much a
statement contributes to the read-out's $\sigma$. Letting a statement float freely
within its tolerance would instead discard the prior's precision exactly where the
fit is weakest.

\paragraph{Gravity as an anchor.}
Because $g$ is a parameter of the ballistic and pendulum laws
(Eq.~\eqref{eq:ballistic}), a free-flight or swinging segment ties lengths to
seconds on its own. Unless the prior states $g$, it is the convention of the
environment, and it becomes one anchor of the metric scale beside the prior's
statements, image measurements, and class votes.

\paragraph{Search.}
The fit builds the scene in the planned order, the camera with the first body,
then each further body, then all of them jointly. Each fitted object then passes
or fails gates. The gates check that its rows hold, its relations hold, and its rendered pose agrees
with its masks within tolerances set by the mask noise and the object's size (Appendix~\ref{app:fit}).

\paragraph{Identifiability.}
\label{sec:method-ident}
A fit returns a value for every free parameter, including those the data do not
constrain, so the fit must also say which values it can vouch for. Let $J$ stack
the Jacobians of the whitened silhouette rows and class votes, $A$ the Jacobian of
the held rows, and $N$ a basis of the null space of $A$, the directions the held
rows leave free. The linearized covariance is
\begin{equation}
  \label{eq:covariance}
  C=s^2\,N\big(N^\top J^\top J N\big)^{+}N^\top ,
\end{equation}
with $s^2$ the residual variance per row. Because a linearization can be
optimistic, a parameter's $\sigma$ is taken conservatively over this and other
estimates that do not rely on it. A parameter is \emph{undetermined}, with a
reason, if its $\sigma$ is large relative to its natural scale or if the data do
not constrain it at all (Appendix~\ref{app:ident}, where Table~\ref{tab:ident} shows one question's
parameters). The metric scale gets its own
test. Scaling all unanchored lengths, speeds, and accelerations together leaves
the image unchanged by construction, so the scale is determined only if some
anchor resists, and testing each anchor alone records whether the scale rests on
the prior, on $g$, on both, or only on class votes.

\paragraph{The read-out.}
The asked quantity is evaluated on the program, a size or $g$ directly and a
speed, acceleration, distance, or height from the laws at the asked time or
window. Its $\sigma$ propagates $C$ through the quantity's gradient, together with
the precision of the statements it rests on and the misfit of the frames it is
read from. The \emph{read-out} is this value only if four conditions hold and the
fitted scene satisfies the prior. Nothing it reads may be open for a structural
reason along which its value changes, such as a direction the data do not curve,
the global scale, or a parameter without observation. Its value must agree within
its $\sigma$ over the equally good states of the fit. Its own uncertainty from the
observations must be at most a quarter of its natural scale, the same bound as for
parameters (Table~\ref{tab:constants}). It must cover its span with a law rather
than a pose table. A parameter that is undetermined only by its own $\sigma$ does
not empty the read-out, since the read-out's $\sigma$ is judged on the read-out.
Otherwise the read-out is empty and records the reasons
(Appendix~\ref{app:ident}). This is the sufficiency test of
Section~\ref{sec:method-msr}, made concrete.
A question may ask about a time at which the queried body is no longer observed.
The read-out then continues the body's fitted motion law to that time and marks
the value as extrapolated. It is withheld when the law stops holding before that
time, as when a free body lands on a support.

\subsection{Answer}
\label{sec:method-answer}

The fit measures magnitudes for a fixed structure, but the structure itself can be
wrong, for example by a missed stop, a wrong support, or a wrong depth group.
Section~\ref{sec:method-fit} already exposes such errors as numbers, residuals that
drift, gates that fail, rows that are violated, and a read-out that is
empty. The method does not revise the structure after the fit. These
numbers stay with the scene, and the answer reads them as evidence.

\paragraph{Read-out and answer.}
Each question receives a read-out and an answer. The read-out of
Section~\ref{sec:method-fit} is empty when the scene does not determine the asked
quantity. The \emph{final answer} comes from one VLM call. It
sees the question, the prior, frames sampled uniformly over the video with their
times, and renderings of the program over the video. It also reads the
reconstruction as evidence. This covers each object's sizes with the source that
gave them, the laws, $g$, the camera, what holds the scale, each prior statement
beside the value the reconstruction has for it, and the reconstruction's value for
the asked quantity in the asked unit. The call is told that the reconstruction
can be right or wrong, and it is not told whether the scene determines the
answer, nor any $\sigma$. It judges from the images whether the reconstruction is
credible for the question and always returns one number. The read-out thus
states, independently of the answer, whether the reconstructed scene measures the
asked quantity.

\section{Experiments}
\label{sec:experiments}

The experiments ask four questions. Does a measured scene give better numbers
than asking a model directly (Section~\ref{sec:exp-quantitative})? Do the
reconstructions agree with the video, also outside the benchmark
(Section~\ref{sec:exp-qualitative})? Does the method know when it has measured
the answer (Section~\ref{sec:exp-reconstruction})? And what do the scene and
the final call each add to the answer (Section~\ref{sec:exp-ablations})?

\subsection{Setup}
\label{sec:exp-setup}

\paragraph{Data.}
We evaluate on the validation split of QuantiPhy~\cite{quantiphy2025}, a
benchmark of quantitative physical questions over simulated, lab-captured, and
internet videos. Each question asks for one number, such as an object's size,
speed, or acceleration, and comes with a prior that fixes the metric scale. The
questions fall into four categories along two axes. The first is the kind of
prior. \emph{Static} (S) priors state a size, and \emph{dynamic} (D) priors state
a speed or an acceleration. The second is whether depth enters. In 2D questions
the prior gives no distance from the camera, and in 3D questions it also gives the
distances from the camera to several named objects. We score the 158 questions
whose ground truth the benchmark releases. They comprise 32 S2, 37 D2, 42 S3, and 47 D3.
\model answers all of them (Appendix~\ref{app:exp}).

\paragraph{Metric.}
We use the benchmark's Mean Relative Accuracy (MRA), the fraction of a fixed set
of relative-error thresholds that a prediction meets, with a missing answer
scoring zero (Appendix~\ref{app:exp}). We report the mean MRA in each category
and, as the overall score, the mean of the four category means, so that each
category weighs equally regardless of its size.

\paragraph{Protocol.}
\model is run once per question, with no repeated sampling and no selection
among runs. The reported answer is the final answer of
Section~\ref{sec:method-answer}, which is always a number. How often the read-out of the
same run is determined is reported in Section~\ref{sec:exp-reconstruction}. No decision of the method reads a
ground-truth answer.

\paragraph{Baselines.}
We re-run every baseline on the same 158 questions. The first is \emph{Code as
Worlds}~\cite{codeasworld} (4B and 9B). The second is \emph{PhysMind}~\cite{physmind}, which
measures the asked quantity on posed metric meshes. The third is \emph{Thinking with Spatial
Code}~\cite{spatialcode2026}, in which a language model reasons over one 3D box per object and the
objects' tracked positions. The last two answer by direct inference.
\emph{Gemini-4-Carbon}~\cite{gemini4carbon} and \emph{Qwen3.5-27B}~\cite{qwen35}
both receive the whole video with the prior and the question. Their settings are in Appendix~\ref{app:exp}.

\subsection{Implementation Details}
\label{sec:exp-implementation}

Every model call of \model goes to one VLM, GPT-6.1~\cite{gpt61sol} (\texttt{gpt-6.1-sol}),
through an OpenAI-compatible endpoint at temperature 0 with a fixed seed. It
writes the object inventory, chooses the tracks, writes and reviews the plan,
authors the shapes, and gives the final answer. Detection, tracking, and monocular
depth use Grounding DINO~\cite{groundingdino2024}, SAM~2~\cite{sam2_2025}, and
Depth Pro~\cite{depthpro2025}.

The plan is written from a grid of nine labelled frames, the fit uses at most 36
frames per object, and the final call sees 16 frames at a long side of at most
768\,px. The fit minimizes its objective with a trust-region least-squares solver.
Table~\ref{tab:constants} lists every constant. All runs use one machine with
128 CPU cores and three NVIDIA H100 NVL GPUs. Table~\ref{tab:timing} gives the
time of each step, and Appendix~\ref{app:exp} gives the model calls per question.

\begin{figure*}[tp]
  \centering
  \includegraphics[width=\linewidth]{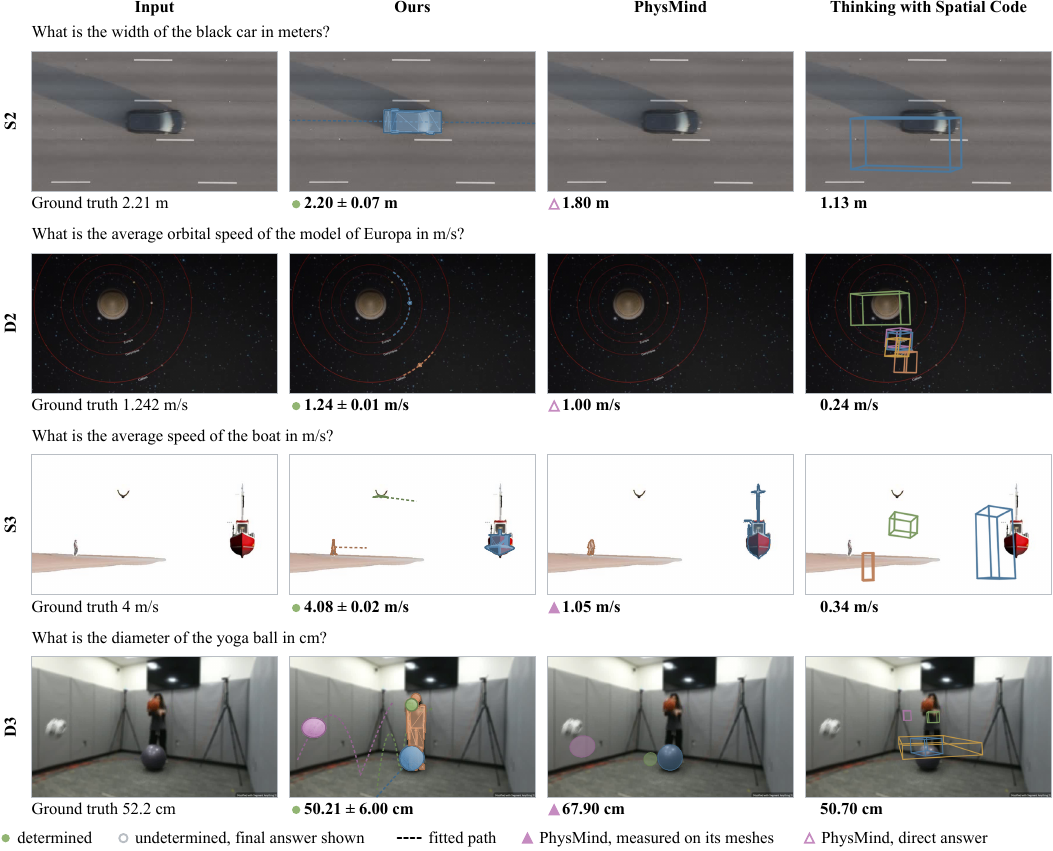}
  \caption{\textbf{Qualitative comparison.} One question per category (rows), each on one frame that every column shows. The question stands above the frames and the answers below them, with the ground truth under the Input column. Each question also comes with its prior, which is not printed. Within a row every object keeps one colour in all columns, and the asked object is blue. The Ours column shows the fitted scene program rendered over the frame, each body with its fitted path dashed, and the read-out with its $\sigma$ (filled dot) or, where the scene leaves the quantity undetermined, the final answer (hollow dot). The PhysMind column shows its posed meshes and its answer, a filled triangle where it was measured on the meshes. Where it produced no meshes, the frame is plain and the answer is the model's direct one (hollow triangle). The Thinking with Spatial Code column shows its 3D boxes and its answer.}
  \Description{The figure has four rows, one per question category. Each row shows the question above the input frame with the ground truth below it, the scene program rendered over the frame with its answer, PhysMind's posed meshes over the frame with its answer, and the 3D boxes of Thinking with Spatial Code over the frame with its answer.}
  \label{fig:qualitative}
\end{figure*}

\subsection{Quantitative Comparison}
\label{sec:exp-quantitative}

Table~\ref{tab:quantitative} compares the final answer of \model with the
baselines, per category and overall. Like theirs, it is a number on every question.

The baselines span the alternatives to a measured scene. Direct inference on the
whole video is the strongest of them. Gemini-4-Carbon reaches 0.624 overall and
leads the baselines on S2, D2, and S3. Qwen3.5-27B, given the same video, reaches
0.521. Questions a model leaves unanswered score zero, 11 for Gemini-4-Carbon and
1 for Qwen3.5-27B (Appendix~\ref{app:exp}). Code as Worlds, which writes the scene as code for
a model to reason over, reaches 0.554 with its larger model and is the strongest
baseline on D3 with 0.581. PhysMind, the
one baseline that measures the answer on a reconstruction, reaches
0.423. Its scale comes from monocular metric depth rather than from the prior, and
on 68 of the 158 questions its own pipeline falls back to the model's direct
answer (Appendix~\ref{app:exp}). Thinking with Spatial Code is the lowest at
0.296. Its code records where each object is in each frame, but ties neither the
positions to a law of motion nor the scale to the prior, which reaches only the
language model that reads the code.

The four categories test different parts of \model. In S2, the stated size
must carry the scale to another object through the camera and the scene's
geometry alone. In D2, the stated speed or acceleration enters as a held row on a
law, and a free-flight segment adds $g$ as a second anchor. In S3 and D3, the given
distances become held rows on the depth of named objects, which fix the camera's
geometry as well as the scale. Over all categories the final answer reaches
0.709, the same when averaged over questions, against 0.624 for the strongest
baseline, Gemini-4-Carbon. By category, the final answer reaches
0.644 on S2, 0.803 on D2, 0.712 on S3, and 0.679 on D3. It leads every baseline
in every category. The margin over the strongest baseline of each category is
smallest on S3, 0.038 over Gemini-4-Carbon, and largest on D2, 0.144 over
Gemini-4-Carbon. It is 0.047 over Gemini-4-Carbon on S2 and 0.098 over Code as
Worlds (9B) on D3.

\begin{table}[t]
  \centering
  \caption{\textbf{Quantitative comparison} on the QuantiPhy validation split
  (MRA, higher is better). S / D denote that the prior states a size, or a speed or
  acceleration. The digits 2 / 3 denote questions without or with given distances from the camera. Overall
  is the mean of the four category means. The authors of Code as Worlds report
  overall scores of 0.506 (4B) and 0.554 (9B). The deeper blue marks the best and the
  lighter blue the second best in each column.}
  \label{tab:quantitative}
  \small
  \setlength{\tabcolsep}{4pt}
  \begin{tabular*}{\columnwidth}{@{\extracolsep{\fill}}lccccc@{}}
    \toprule
    Method & S2 & D2 & S3 & D3 & Overall \\
    Questions & 32 & 37 & 42 & 47 & 158 \\
    \midrule
    Code as Worlds (4B) & 0.409 & 0.557 & 0.455 & 0.523 & 0.486 \\
    Code as Worlds (9B) & 0.559 & 0.519 & 0.557 & \second{0.581} & 0.554 \\
    PhysMind            & 0.300 & 0.508 & 0.412 & 0.470 & 0.423 \\
    \begin{tabular}[t]{@{}l@{}}Thinking with\\\hspace{0.8em}Spatial Code\end{tabular} & 0.222 & 0.257 & 0.352 & 0.353 & 0.296 \\
    Gemini-4-Carbon     & \second{0.597} & \second{0.659} & \second{0.674} & 0.566 & \second{0.624} \\
    Qwen3.5-27B         & 0.512 & 0.489 & 0.536 & 0.545 & 0.521 \\
    \midrule
    \model & \best{0.644} & \best{0.803} & \best{0.712} & \best{0.679} & \best{0.709} \\
    \bottomrule
  \end{tabular*}
\end{table}

\begin{figure*}[tp]
  \centering
  \includegraphics[width=\linewidth]{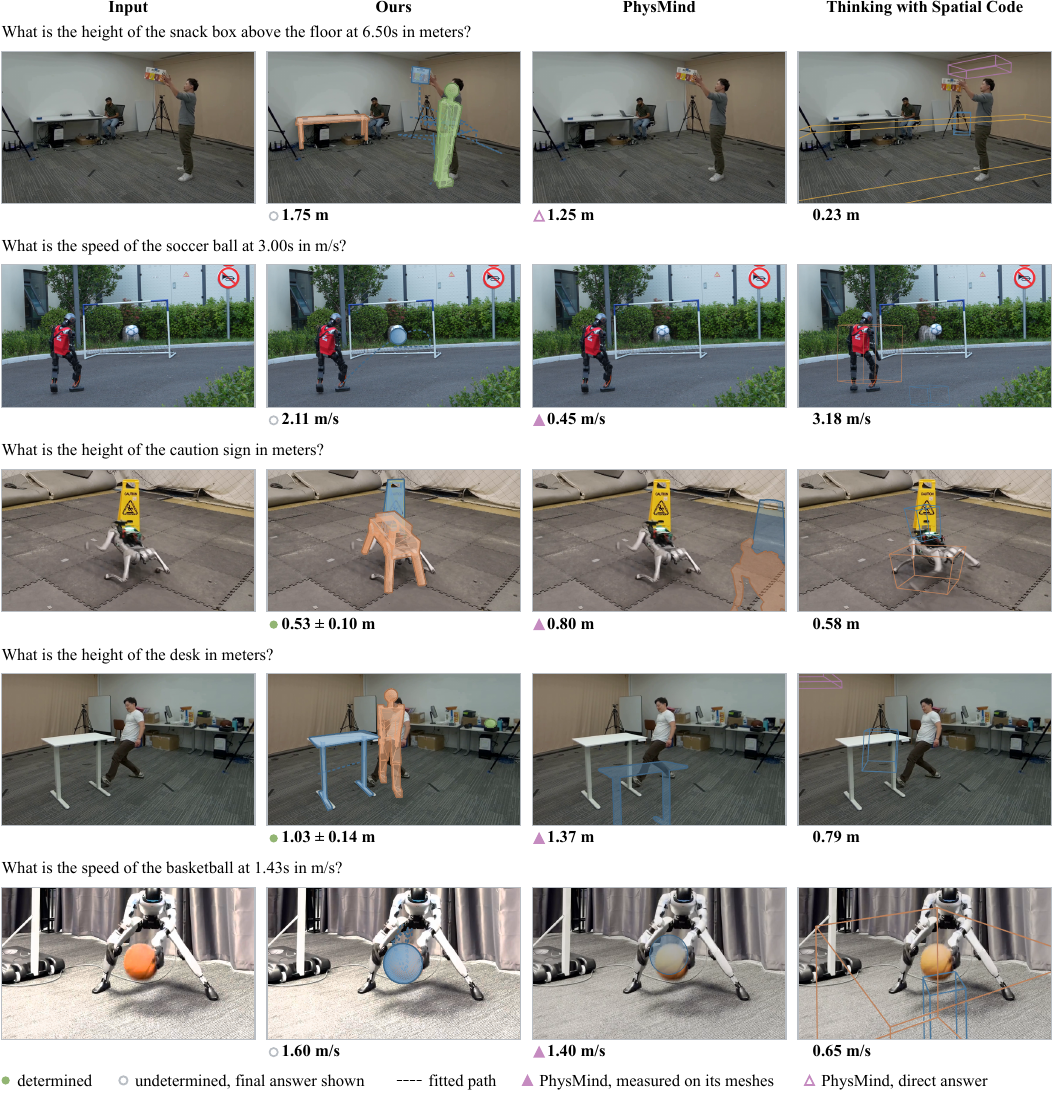}
  \caption{\textbf{In-the-wild videos.} Five questions on videos outside the benchmark. These videos have no ground truth, so the figure shows each method's reconstruction and answer without ranking them. Each row has the question above the frames and the answers below them, on the frame on which PhysMind and Thinking with Spatial Code drew their reconstructions. Within a row every object keeps one colour in all columns, and the asked object is blue. The Ours column shows the fitted scene program with the asked body's fitted path dashed, and the read-out with its $\sigma$ (filled dot) or, where the scene leaves the quantity undetermined, the final answer (hollow dot). The PhysMind column shows its posed meshes and its answer, a filled triangle where it was measured on the meshes and a hollow triangle where its pipeline fell back to the model's direct answer. The Thinking with Spatial Code column shows its 3D boxes and its answer. Appendix~\ref{app:wild} describes the videos.}
  \Description{Five rows, one per in-the-wild question, from lab recordings, robot soccer and basketball, and a quadruped robot. Each row shows the question above four frames, the input, our scene program rendered over it, PhysMind's posed meshes, and the 3D boxes of Thinking with Spatial Code, with each method's answer below its frame.}
  \label{fig:wild}
\end{figure*}

\subsection{Qualitative Results}
\label{sec:exp-qualitative}

A number can be right for the wrong reason, so we also inspect the scenes the
numbers are read from. Figure~\ref{fig:qualitative} shows, for one question of
each category, the input frame, the scene program rendered over it, and the
reconstructions of PhysMind and of Thinking with Spatial Code.
PhysMind is the baseline whose scene is closest to ours. The
two reconstructions make opposite trades. PhysMind's meshes carry detailed shape,
but their placement and scale come from per-frame depth and pose estimates, so
nothing ties an object's motion across frames to a law or its size to the prior.
\model's shapes are coarse, but each pose is the value of a fitted law and
the scale is held by the prior. What to look for is whether the rendered program
stays on the silhouettes over the whole video, and whether the parameter the
question reads is marked determined. Thinking with Spatial Code sits at the other
extreme, with one box per object, whose size and placement come from its encoder.
Where a baseline produces a
reconstruction, it mostly finds the objects but gets the metric scale wrong. On the
highway (S2), PhysMind produced no meshes and answers 0.81 times the truth. The box
of Thinking with Spatial Code trails the car and is about half as wide. The program's
car sits on its silhouette, the given lane width holds its scale, and its read-out is
determined and within 1\% of the truth. Around the planet model (D2), the prior is the speed of the Callisto
model. PhysMind produced no meshes and answers 0.81 times the truth. The boxes of Thinking
with Spatial Code gather below the planet, away from the moons, and its speed is a fifth
of the truth. In the program both moons follow circular laws in one plane, which carries
the stated speed to Europa, and its read-out is determined and within 1\%. The boat (S3) is more than 90\,m from the camera. PhysMind's mesh and
the box of Thinking with Spatial Code both cover it, but their speeds are a quarter
of the truth or less, while the given distances hold the program's scale and its
speed is within 3\%. In the room (D3), PhysMind's mesh sits on the yoga ball yet is 1.3
times too large. The box of Thinking with Spatial Code is offset from the ball, and
its answer lies within 3\% of the truth. The program places the ball on the
fitted floor, and its read-out is determined and within 4\%.

\paragraph{In-the-wild videos.}
To see whether the method depends on the benchmark's videos, we apply \model
unchanged to 27 real videos from outside it (Figure~\ref{fig:wild}). They show robot
soccer and basketball, quadruped robots, and lab recordings of objects that are
pushed or tossed. These videos have no ground-truth answers, and we do not score
them. On each video we pose two quantitative questions about a named
object, its size, its speed at a given time, or a distance or height, written
from the frames before any run. Each video's prior is one standard public fact
that a user would know without measuring. It is the regulation diameter of the ball
(0.22\,m for a size-5 football, 0.24\,m for a size-7 basketball), the
manufacturer's published body length of the quadruped robot, or the 1.22\,m
length of a standard 2\,$\times$\,4\,ft ceiling light panel. Every video has a
static camera and one object of standard size (Appendix~\ref{app:wild}).
What they test is alignment, namely whether the program, rendered back onto each frame,
follows the bodies through contact, rolling, and flight, and whether the fit
reports the scale and the asked quantity as determined. Tossed objects add a
second anchor, since a ballistic segment reads $g$, and the scale test records
whether the scale rests on the stated size, on $g$, or on both. Of the 54
questions, 45 on 24 of the videos were answered (Appendix~\ref{app:wild}). On all
45 the scale is determined, by the stated size alone on 28 and together with $g$
on 17.

\subsection{Reconstruction Quality}
\label{sec:exp-reconstruction}

The fit reports on its own scene without answer labels, so the reconstructions
can be judged independently of accuracy, and the method's claim that it knows
when it has measured the answer can be tested against the labels afterwards. We
read two fractions from these reports. The first is the fraction of questions
whose metric scale is determined, all 158, and which anchors it rests
on (Table~\ref{tab:anchors}). The second is the fraction whose read-out is
determined. It is 47 of the 158 questions (0.297), with 6 of 32 in S2, 15 of 37 in D2,
14 of 42 in S3, and 12 of 47 in D3.

The determined flag is useful only if it separates good measurements from bad
ones. If it does, the final answer should be more accurate where it rests on a
determined read-out than where it does not. It is. On the 47 determined
questions the final answer reaches 0.781, against 0.679 on the 111 undetermined
ones.

Table~\ref{tab:anchors} groups the questions by what their scale rests on and
gives the MRA of the final answer in each group. The prior holds the scale on 127
questions, and the prior together with $g$, in a law the plan writes with gravity,
on 31.

\begin{table}[t]
  \centering
  \caption{\textbf{Scale anchors.} Questions grouped by what the scale test finds
  the metric scale rests on, with the MRA of the final answer in each group. An image
  measurement binds a prior statement and is counted under the prior, and so is
  a $g$ the prior states. Gravity $g$ counts as an anchor of its own only where the
  program takes it as the environment's convention. Read from the scale test of
  the fit (Appendix~\ref{app:exp}).}
  \label{tab:anchors}
  \small
  \begin{tabular*}{\columnwidth}{@{\extracolsep{\fill}}lcc@{}}
    \toprule
    Scale rests on & Questions & MRA \\
    \midrule
    The prior only      & 127 & 0.720 \\
    The prior and $g$   & 31  & 0.668 \\
    \bottomrule
  \end{tabular*}
\end{table}

\subsection{Where the Final Answer Comes From}
\label{sec:exp-ablations}

The final call sees the reconstruction's value for the question as evidence, but
not its $\sigma$ or whether the read-out is determined. Splitting the questions by
this flag separates what the scene measures from what the final call adds. On the
47 questions with a determined read-out, the final answer reaches 0.781, against
0.745 for the scene value itself. The final call stays within 1\% of the scene
value on 36 of them, at an MRA of 0.794, and writes a number of its own on 11, at
0.736. On the 111 undetermined questions it reaches 0.679. The read-out offers a
value it does not vouch for on 100 of them. The final call stays within 1\% of
that value on 35, at 0.669, and writes its own number on 65, at 0.677. The 11
questions without a scene value reach 0.727. Of the 24 questions whose final
answer scores below 0.5, 3 keep a determined scene value to within 1\%, and
Appendix~\ref{app:failures} traces each of the 24 to the earliest stage at which
it goes wrong.

\section{Conclusion}
\label{sec:conclusion}

\model answers quantitative questions about video by reconstructing, for each
question, the smallest scene that suffices to measure the asked quantity, and
answering with that scene as evidence. The scene is a program whose structure a VLM writes
and whose magnitudes only the prior, conventions, class sizes, and the fit may
set. Physics-guided Fitting measures those magnitudes through motion laws, so
that a few physical parameters absorb many noisy frames and gravity can anchor
the scale beside the prior, and it states which parameters, and whether the
read-out of the asked quantity, the evidence determines. The final answer reads that scene as evidence
beside the frames. What this buys is an answer that can be inspected. Every number traces to the objects,
scale, and motion behind it, and the read-out says on its own whether the
reconstruction measured what was asked. The limitations of the approach are
discussed in Appendix~\ref{app:limitations}.

Three directions follow. The first is more data. The in-the-wild videos show that the method runs outside the benchmark, but they cannot score it, so labelled datasets beyond QuantiPhy are needed to validate it. The second is moving cameras, which the scene program could treat as
one more body with its own motion law, anchored by the static background. The
third is finer shapes, refined only where a coarse solid leaves the asked
quantity undetermined, in keeping with the principle of reconstructing no more
than the question needs.

\bibliographystyle{ACM-Reference-Format}
\bibliography{refs}

@inproceedings{clevrer2020,
  title     = {{CLEVRER}: Collision Events for Video Representation and Reasoning},
  author    = {Yi, Kexin and Gan, Chuang and Li, Yunzhu and Kohli, Pushmeet and Wu, Jiajun and Torralba, Antonio and Tenenbaum, Joshua B.},
  booktitle = {International Conference on Learning Representations (ICLR)},
  year      = {2020}
}

@inproceedings{physion2021,
  title     = {{Physion}: Evaluating Physical Prediction from Vision in Humans and Machines},
  author    = {Bear, Daniel and Wang, Elias and Mrowca, Damian and Binder, Felix J. and Tung, Hsiao-Yu and Pramod, R. T. and Holdaway, Cameron and Tao, Sirui and Smith, Kevin A. and Sun, Fan-Yun and Fei-Fei, Li and Kanwisher, Nancy and Tenenbaum, Joshua B. and Yamins, Dan and Fan, Judith E.},
  booktitle = {Advances in Neural Information Processing Systems (NeurIPS), Datasets and Benchmarks Track},
  year      = {2021}
}

@inproceedings{physionplus2023,
  title     = {{Physion++}: Evaluating Physical Scene Understanding that Requires Online Inference of Different Physical Properties},
  author    = {Tung, Hsiao-Yu and Ding, Mingyu and Chen, Zhenfang and Bear, Daniel and Gan, Chuang and Tenenbaum, Joshua B. and Yamins, Daniel LK and Fan, Judith E. and Smith, Kevin A.},
  booktitle = {Advances in Neural Information Processing Systems (NeurIPS), Datasets and Benchmarks Track},
  year      = {2023}
}

@inproceedings{physbench2024,
  title     = {{PhysBench}: Benchmarking and Enhancing Vision-Language Models for Physical World Understanding},
  author    = {Chow, Wei and Mao, Jiageng and Li, Boyi and Seita, Daniel and Guizilini, Vitor Campagnolo and Wang, Yue},
  booktitle = {International Conference on Learning Representations (ICLR)},
  year      = {2025}
}

@inproceedings{quantiphy2025,
  title     = {{QuantiPhy}: A Quantitative Benchmark Evaluating Physical Reasoning Abilities of Vision-Language Models},
  author    = {Puyin, Li and Xiang, Tiange and Mao, Ella and Wei, Shirley and Chen, Xinye and Masood, Adnan and Fei-Fei, Li and Adeli, Ehsan},
  booktitle = {{IEEE/CVF} Conference on Computer Vision and Pattern Recognition (CVPR)},
  year      = {2026}
}

@inproceedings{dpt2021,
  title     = {Vision Transformers for Dense Prediction},
  author    = {Ranftl, Ren{\'e} and Bochkovskiy, Alexey and Koltun, Vladlen},
  booktitle = {{IEEE/CVF} International Conference on Computer Vision (ICCV)},
  year      = {2021}
}

@inproceedings{depthanything2024,
  title     = {{Depth Anything}: Unleashing the Power of Large-Scale Unlabeled Data},
  author    = {Yang, Lihe and Kang, Bingyi and Huang, Zilong and Xu, Xiaogang and Feng, Jiashi and Zhao, Hengshuang},
  booktitle = {{IEEE/CVF} Conference on Computer Vision and Pattern Recognition (CVPR)},
  year      = {2024}
}

@article{zoedepth2023,
  title   = {{ZoeDepth}: Zero-shot Transfer by Combining Relative and Metric Depth},
  author  = {Bhat, Shariq Farooq and Birkl, Reiner and Wofk, Diana and Wonka, Peter and M{\"u}ller, Matthias},
  journal = {arXiv preprint arXiv:2302.12288},
  year    = {2023}
}

@inproceedings{cotracker2023,
  title     = {{CoTracker}: It Is Better to Track Together},
  author    = {Karaev, Nikita and Rocco, Ignacio and Graham, Benjamin and Neverova, Natalia and Vedaldi, Andrea and Rupprecht, Christian},
  booktitle = {European Conference on Computer Vision (ECCV)},
  year      = {2024}
}

@inproceedings{dust3r2024,
  title     = {{DUSt3R}: Geometric {3D} Vision Made Easy},
  author    = {Wang, Shuzhe and Leroy, Vincent and Cabon, Yohann and Chidlovskii, Boris and Revaud, J{\'e}r{\^o}me},
  booktitle = {{IEEE/CVF} Conference on Computer Vision and Pattern Recognition (CVPR)},
  year      = {2024}
}

@inproceedings{npe2017,
  title     = {A Compositional Object-Based Approach to Learning Physical Dynamics},
  author    = {Chang, Michael and Ullman, Tomer D. and Torralba, Antonio and Tenenbaum, Joshua B.},
  booktitle = {International Conference on Learning Representations (ICLR)},
  year      = {2017}
}

@inproceedings{phydnet2020,
  title     = {Disentangling Physical Dynamics from Unknown Factors for Unsupervised Video Prediction},
  author    = {Le Guen, Vincent and Thome, Nicolas},
  booktitle = {{IEEE/CVF} Conference on Computer Vision and Pattern Recognition (CVPR)},
  year      = {2020}
}

@inproceedings{visprog2023,
  title     = {Visual Programming: Compositional Visual Reasoning without Training},
  author    = {Gupta, Tanmay and Kembhavi, Aniruddha},
  booktitle = {{IEEE/CVF} Conference on Computer Vision and Pattern Recognition (CVPR)},
  year      = {2023}
}

@inproceedings{vipergpt2023,
  title     = {{ViperGPT}: Visual Inference via {Python} Execution for Reasoning},
  author    = {Sur{\'\i}s, D{\'\i}dac and Menon, Sachit and Vondrick, Carl},
  booktitle = {{IEEE/CVF} International Conference on Computer Vision (ICCV)},
  year      = {2023}
}

@inproceedings{depthpro2025,
  title     = {{Depth Pro}: Sharp Monocular Metric Depth in Less Than a Second},
  author    = {Bochkovskii, Aleksei and Delaunoy, Ama{\"e}l and Germain, Hugo and Santos, Marcel and Zhou, Yichao and Richter, Stephan R. and Koltun, Vladlen},
  booktitle = {International Conference on Learning Representations (ICLR)},
  year      = {2025}
}

@inproceedings{monst3r2025,
  title     = {{MonST3R}: A Simple Approach for Estimating Geometry in the Presence of Motion},
  author    = {Zhang, Junyi and Herrmann, Charles and Hur, Junhwa and Jampani, Varun and Darrell, Trevor and Cole, Forrester and Sun, Deqing and Yang, Ming-Hsuan},
  booktitle = {International Conference on Learning Representations (ICLR)},
  year      = {2025}
}

@inproceedings{megasam2025,
  title     = {{MegaSaM}: Accurate, Fast and Robust Structure and Motion from Casual Dynamic Videos},
  author    = {Li, Zhengqi and Tucker, Richard and Cole, Forrester and Wang, Qianqian and Jin, Linyi and Ye, Vickie and Kanazawa, Angjoo and Holynski, Aleksander and Snavely, Noah},
  booktitle = {{IEEE/CVF} Conference on Computer Vision and Pattern Recognition (CVPR)},
  year      = {2025}
}

@article{codeasworld,
  title     = {Code as Worlds: Agentic Discovery of Executable World Representations for Physical Reasoning},
  author    = {Wang, Hanyang and Cai, Yimo and Chen, Weiliang and Chi, Jiawei and Sun, Haowen and Dai, Qiyu and Hung, Yi{-}Hsin and Guo, Xingzhuo and Ren, Jinshan and Yao, Runmao and Liu, Ziwei and Long, Mingsheng and Duan, Yueqi and Gao, Jun and Lyu, Jiangran and Liu, Fangfu and Wu, Jialong},
  journal   = {arXiv preprint arXiv:2608.27549},
  year      = {2026}
}

@article{physmind,
  title     = {{PhysMind}: From Video to Executable Worlds for Training-Free Physical Reasoning},
  author    = {Yang, Chen and Zeng, Shenxiang and Zhao, Haoyang and Xu, Zhouyuan and He, Youquan and Li, Haoyu and Deng, Mingyi and Fan, Jiansheng and Wang, Chen},
  journal   = {arXiv preprint arXiv:2608.04575},
  year      = {2026}
}

@inproceedings{spatialvlm2024,
  title     = {{SpatialVLM}: Endowing Vision-Language Models with Spatial Reasoning Capabilities},
  author    = {Chen, Boyuan and Xu, Zhuo and Kirmani, Sean and Ichter, Brian and Sadigh, Dorsa and Guibas, Leonidas J. and Xia, Fei},
  booktitle = {{IEEE/CVF} Conference on Computer Vision and Pattern Recognition (CVPR)},
  year      = {2024}
}

@inproceedings{spatialrgpt2024,
  title     = {{SpatialRGPT}: Grounded Spatial Reasoning in Vision-Language Models},
  author    = {Cheng, An{-}Chieh and Yin, Hongxu and Fu, Yang and Guo, Qiushan and Yang, Ruihan and Kautz, Jan and Wang, Xiaolong and Liu, Sifei},
  booktitle = {Advances in Neural Information Processing Systems (NeurIPS)},
  year      = {2024}
}

@inproceedings{moge2,
  title     = {{MoGe-2}: Accurate Monocular Geometry with Metric Scale and Sharp Details},
  author    = {Wang, Ruicheng and Xu, Sicheng and Dong, Yue and Deng, Yu and Xiang, Jianfeng and Lv, Zelong and Sun, Guangzhong and Tong, Xin and Yang, Jiaolong},
  booktitle = {Advances in Neural Information Processing Systems (NeurIPS)},
  year      = {2025}
}

@inproceedings{sam3dobjects,
  author    = {Chen, Xingyu and Chu, Fu{-}Jen and Gleize, Pierre and Liang, Kevin J. and Sax, Alexander and Tang, Hao and Wang, Weiyao and Guo, Michelle and Hardin, Thibaut and Li, Xiang and Lin, Aohan and Liu, Jia{-}Wei and Ma, Ziqi and Sagar, Anushka and Song, Bowen and Wang, Xiaodong and Yang, Jianing and Zhang, Bowen and Doll{\'{a}}r, Piotr and Gkioxari, Georgia and Feiszli, Matt and Malik, Jitendra},
  title     = {{SAM 3D}: {3D}fy Anything in Images},
  booktitle = {{IEEE/CVF} Conference on Computer Vision and Pattern Recognition (CVPR)},
  year      = {2026}
}

@inproceedings{foundationpose2024,
  title     = {{FoundationPose}: Unified {6D} Pose Estimation and Tracking of Novel Objects},
  author    = {Wen, Bowen and Yang, Wei and Kautz, Jan and Birchfield, Stan},
  booktitle = {{IEEE/CVF} Conference on Computer Vision and Pattern Recognition (CVPR)},
  year      = {2024}
}

@inproceedings{groundingdino2024,
  title     = {Grounding {DINO:} Marrying {DINO} with Grounded Pre-training for Open-Set Object Detection},
  author    = {Liu, Shilong and Zeng, Zhaoyang and Ren, Tianhe and Li, Feng and Zhang, Hao and Yang, Jie and Jiang, Qing and Li, Chunyuan and Yang, Jianwei and Su, Hang and Zhu, Jun and Zhang, Lei},
  booktitle = {European Conference on Computer Vision (ECCV)},
  year      = {2024}
}

@inproceedings{sam2_2025,
  author    = {Ravi, Nikhila and Gabeur, Valentin and Hu, Yuan{-}Ting and Hu, Ronghang and Ryali, Chaitanya and Ma, Tengyu and Khedr, Haitham and R{\"{a}}dle, Roman and Rolland, Chlo{\'{e}} and Gustafson, Laura and Mintun, Eric and Pan, Junting and Alwala, Kalyan Vasudev and Carion, Nicolas and Wu, Chao{-}Yuan and Girshick, Ross B. and Doll{\'{a}}r, Piotr and Feichtenhofer, Christoph},
  title     = {{SAM} 2: Segment Anything in Images and Videos},
  booktitle = {International Conference on Learning Representations (ICLR)},
  year      = {2025}
}

@misc{qwen35,
  author       = {{Qwen Team}},
  title        = {{Qwen3.5}: Towards Native Multimodal Agents},
  year         = {2026},
  month        = feb,
  howpublished = {\url{https://qwen.ai/blog?id=qwen3.5}},
  note         = {Accessed: 2026-11-13}
}

@inproceedings{galileo2015,
  title     = {{Galileo}: Perceiving Physical Object Properties by Integrating a Physics Engine with Deep Learning},
  author    = {Wu, Jiajun and Yildirim, Ilker and Lim, Joseph J. and Freeman, William T. and Tenenbaum, Joshua B.},
  booktitle = {Advances in Neural Information Processing Systems (NIPS)},
  year      = {2015}
}

@inproceedings{jaques2020physics,
  title     = {Physics-as-Inverse-Graphics: Unsupervised Physical Parameter Estimation from Video},
  author    = {Jaques, Miguel and Burke, Michael and Hospedales, Timothy},
  booktitle = {International Conference on Learning Representations (ICLR)},
  year      = {2020}
}

@article{kandukuri2022physical,
  title   = {Physical Representation Learning and Parameter Identification from Video Using Differentiable Physics},
  author  = {Kandukuri, Rama Krishna and Achterhold, Jan and Moeller, Michael and Stueckler, Joerg},
  journal = {International Journal of Computer Vision},
  year    = {2022}
}

@article{spatialcode2026,
  title   = {Thinking with Spatial Code for Physical-World Video Reasoning},
  author  = {Chen, Jieneng and Ma, Wenxin and Yuan, Ruisheng and Zhang, Yunzhi and Wu, Jiajun and Yuille, Alan},
  journal = {arXiv preprint arXiv:2603.05591},
  year    = {2026}
}

@misc{gpt61sol,
  author       = {{OpenAI}},
  title        = {Introducing {GPT-6.1 Sol}},
  year         = {2026},
  month        = sep,
  howpublished = {\url{https://openai.com/index/introducing-gpt-6-1-sol/}},
  note         = {Accessed: 2026-11-13}
}

@inproceedings{sam3_2026,
  title     = {{SAM} 3: Segment Anything with Concepts},
  author    = {Carion, Nicolas and Gustafson, Laura and Hu, Yuan{-}Ting and Debnath, Shoubhik and Hu, Ronghang and Sur{\'\i}s, D{\'\i}dac and Ryali, Chaitanya and Alwala, Kalyan Vasudev and Khedr, Haitham and Huang, Andrew and Lei, Jie and Ma, Tengyu and Guo, Baishan and Kalla, Arpit and Marks, Markus and Greer, Joseph and Wang, Meng and Sun, Peize and R{\"{a}}dle, Roman and Afouras, Triantafyllos and Mavroudi, Effrosyni and Xu, Katherine and Wu, Tsung{-}Han and Zhou, Yu and Momeni, Liliane and Hazra, Rishi and Ding, Shuangrui and Vaze, Sagar and Porcher, Francois and Li, Feng and Li, Siyuan and Kamath, Aishwarya and Cheng, Ho Kei and Doll{\'{a}}r, Piotr and Ravi, Nikhila and Saenko, Kate and Zhang, Pengchuan and Feichtenhofer, Christoph},
  booktitle = {International Conference on Learning Representations (ICLR)},
  year      = {2026}
}

@inproceedings{videodepthanything2025,
  title     = {{Video Depth Anything}: Consistent Depth Estimation for Super-Long Videos},
  author    = {Chen, Sili and Guo, Hengkai and Zhu, Shengnan and Zhang, Feihu and Huang, Zilong and Feng, Jiashi and Kang, Bingyi},
  booktitle = {{IEEE/CVF} Conference on Computer Vision and Pattern Recognition (CVPR)},
  year      = {2025}
}

@inproceedings{geocalib2024,
  title     = {{GeoCalib}: Learning Single-image Calibration with Geometric Optimization},
  author    = {Veicht, Alexander and Sarlin, Paul{-}Edouard and Lindenberger, Philipp and Pollefeys, Marc},
  booktitle = {European Conference on Computer Vision (ECCV)},
  year      = {2024}
}

@misc{gemini3flash,
  author       = {{Gemini Team, Google}},
  title        = {{Gemini 3 Flash} Model Card},
  year         = {2025},
  month        = dec,
  howpublished = {\url{https://deepmind.google/models/model-cards/gemini-3-flash/}},
  note         = {Accessed: 2026-11-13}
}

@article{qwen3,
  title   = {{Qwen3} Technical Report},
  author  = {{Qwen Team}},
  journal = {arXiv preprint arXiv:2505.09388},
  year    = {2025}
}

@article{drillis1964body,
  title={Body segment parameters},
  author={Drillis, Rudolfs and Contini, Renato and Bluestein, Maurice},
  journal={Artif. limbs},
  volume={8},
  number={1},
  pages={44--66},
  year={1964}
}

@book{winter2009biomechanics,
  title={Biomechanics and motor control of human movement},
  author={Winter, David A.},
  edition={4},
  year={2009},
  publisher= {John Wiley \& Sons},
  address= {Hoboken, NJ}
}

@inproceedings{johnson2015image,
  title={Image retrieval using scene graphs},
  author={Johnson, Justin and Krishna, Ranjay and Stark, Michael and Li, Li-Jia and Shamma, David and Bernstein, Michael and Fei-Fei, Li},
  booktitle={{IEEE} Conference on Computer Vision and Pattern Recognition (CVPR)},
  pages={3668--3678},
  year={2015}
}

@inproceedings{armeni20193d,
  title={{3D} Scene Graph: A structure for unified semantics, {3D} space, and camera},
  author={Armeni, Iro and He, Zhi-Yang and Gwak, JunYoung and Zamir, Amir R and Fischer, Martin and Malik, Jitendra and Savarese, Silvio},
  booktitle={{IEEE/CVF} International Conference on Computer Vision (ICCV)},
  pages={5664--5673},
  year={2019}
}

@inproceedings{rosinol20203d,
  title= {{3D} Dynamic Scene Graphs: Actionable Spatial Perception with Places, Objects, and Humans},
  author= {Rosinol, Antoni and Gupta, Arjun and Abate, Marcus and Shi, Jingnan and Carlone, Luca},
  booktitle = {Robotics: Science and Systems (RSS)},
  year= {2020},
  doi= {10.15607/RSS.2020.XVI.079}
}

@misc{gemini4carbon,
  title= {{Gemini 4 Carbon} Model Card},
  author= {{Google DeepMind}},
  howpublished= {Google DeepMind model card},
  year= {2026},
  url= {https://deepmind.google/models/model-cards/gemini-4-carbon/},
  note= {Accessed: 2026-11-13}
}

\clearpage
\appendix
\newcommand{\apptocline}[3]{\noindent\hspace*{#1}\makebox[2.4em][l]{\ref{#2}}#3\dotfill\pageref{#2}\par}
\noindent{\large\bfseries Contents of the Appendix}\par\medskip
\apptocline{0em}{app:method}{Method Details}
\apptocline{1em}{app:inputs}{Observations and Prior}
\apptocline{1em}{app:plan}{The Plan}
\apptocline{1em}{app:laws}{Objects, Motion Laws, and Camera}
\apptocline{1em}{app:fit}{Fitting}
\apptocline{1em}{app:ident}{Identifiability and the Read-out}
\apptocline{1em}{app:answer}{Answer Step}
\apptocline{0em}{app:exp}{Experimental Details}
\apptocline{1em}{app:wild}{In-the-Wild Videos}
\apptocline{0em}{app:failures}{Failure Cases}
\apptocline{1em}{app:gt-doubt}{Ground Truth in Doubt}
\apptocline{0em}{app:limitations}{Limitations}
\bigskip

\section{Method Details}
\label{app:method}

\begin{figure*}[tp]
  \centering
  \includegraphics[width=\linewidth]{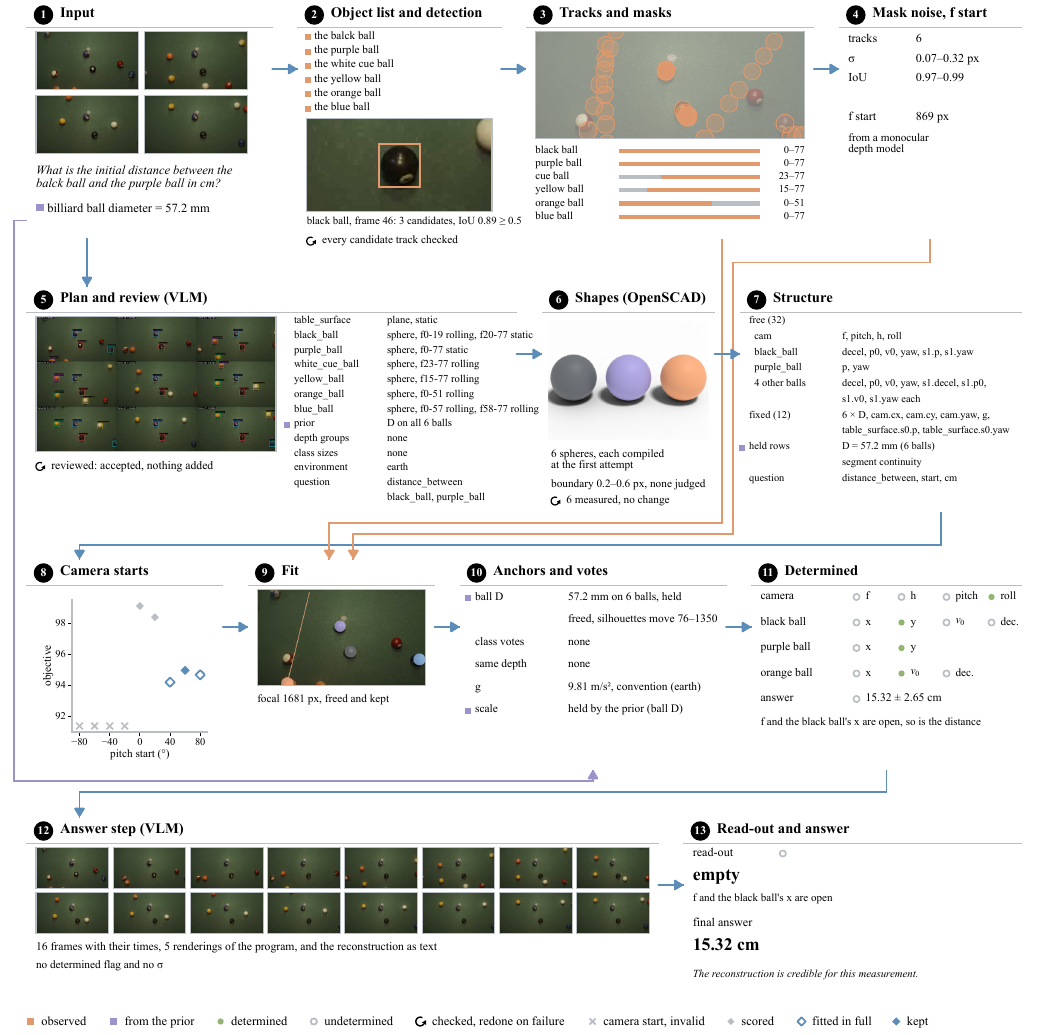}
  \caption{\textbf{Every stage on one question}, the initial distance between the black and the purple billiard ball, with the prior ``billiard ball diameter = 57.2\,mm''. Panels are numbered in reading order and arrows show what each stage passes on. Return arrows mark the stages that check their own product, the candidate tracks (2), the plan (5), and the shapes (6). (1) The input. (2) The object list and the chosen detection. (3) The tracks with their masks and frame ranges. (4) The tracks' mask noise and consecutive-mask IoU, and the focal length the camera starts from. (5) The plan on the labelled frame grid, and its review. (6) The authored shapes, each measured against its masks, none judged. (7) The structure, with free and fixed parameters, held rows, and the parsed question. (8) The camera's pitch starts. A cross means the table would lie above the camera, a hollow diamond means the start is carried through the full fit, and the filled blue diamond is the start kept. (9) The fit over the video, which also receives the masks and the focal start. (10) The anchors and votes. The stated diameter holds the scale. (11) The parameters the fit determines. The focal length, the camera's height and pitch, and the black ball's position along the table are open, so the distance is undetermined. (12) The answer step, which sees 16 timestamped frames, renderings of the program, and the reconstruction as text, without the determined flag or $\sigma$. (13) The read-out and the answer. The read-out is empty, and the final answer is 15.32\,cm.}
  \Description{A four-row grid of panels. Row one shows video frames, question and prior, the track names and a crop with the detected box, mask outlines over a frame with bars for each track's frames, and mask noise ranges. Row two shows a nine-frame grid with the plan as text, three rendered ball shapes, and the structure as text. Row three shows a scatter of objective against pitch start, a fitted overlay, anchors, and filled and hollow dots for parameters. Row four shows the sixteen frames the answer step sees and the scene and final answers.}
  \label{fig:pipeline-full}
\end{figure*}

This appendix gives the details behind Section~\ref{sec:method}, namely the inputs, the
rules on the plan, the motion laws, the mechanics of the fit, the identifiability
tests, and the answer step.
Table~\ref{tab:constants} collects the constants, and
Figure~\ref{fig:pipeline-full} follows the question of Figure~\ref{fig:pipeline}
through every stage.

\begin{table}[t]
  \centering
  \caption{Constants of the method.}
  \label{tab:constants}
  \small
  \begin{tabular*}{\columnwidth}{@{\extracolsep{\fill}}lr@{}}
    \toprule
    \multicolumn{2}{@{}l}{\emph{Plan and shapes}}\\
    Plan grid frames & 9\\
    Shape attempts per body & 3\\
    Shape check rounds per body & 3\\
    Shape check time per question & 120\,s\\
    \midrule
    \multicolumn{2}{@{}l}{\emph{Fit}}\\
    Precision groups per frame $G$ & 32\\
    Recall samples per frame $K$ & $\le 48$\\
    Samples per frame and term $K_{\mathrm{e}}$ & 32\\
    Pull to initial values $\lambda$ & $10^{-3}$\\
    Fitted frames per object & $\le 36$\\
    Pitch starts, every $20^\circ$ & $-80^\circ$--$80^\circ$\\
    Starts fitted in full & $\ge 3$\\
    Diagonal field of view & $1^\circ$--$120^\circ$\\
    Segment boundary shift & $\le 3$ frames\\
    Projection precision $\epsilon_0$ & $10^{-9}$\\
    Pixel floor of thresholds & $\sqrt{2}$\,px\\
    \midrule
    \multicolumn{2}{@{}l}{\emph{Identifiability}}\\
    Determined if $\sigma/$natural scale $\le$ (parameter, read-out) & 0.25\\
    Scale test factor & 1.01\\
    Class vote in tension & $3\sigma_c$\\
    \midrule
    \multicolumn{2}{@{}l}{\emph{Answer}}\\
    Frames for the final answer & 16\\
    Overlays for the final answer & $\le 5$\\
    \bottomrule
  \end{tabular*}
\end{table}

\subsection{Observations and Prior}
\label{app:inputs}

A text-only model call lists the distinct objects that the question and the prior
name, merging different spellings of one object, and may add a visible reference
object, such as one of well-known size or a ball in free flight. Each object is
detected with Grounding DINO, the VLM picks the prompt box among the detector's
candidates, and SAM~2 propagates it through the video. A candidate track is kept
only if it passes named checks of identity, presence, size, and the image border.
An object whose candidates all fail stays untracked. On every tracked frame, SAM~2
segments the object inside its track box, giving one binary mask per object and
frame.

The mask noise $\sigma_o$ is the median boundary distance between an object's
masks on consecutive frames after their centroids are aligned, measured on frames
where the mask is neither cut by the image border nor shared with another track.
Every pixel threshold is stated in units of $\sigma_o$, with a floor of
$\sqrt{2}$\,px for two rasterized boundaries.

The prior is parsed into numbered statements, each with a quantity, an object
phrase, an SI value, and, where given, a time. A distance statement becomes a
statement of the quantity \emph{distance to camera}, the depth of the object's
center along the optical axis. The program never re-detects objects and never
changes a statement's value.

\subsection{The Plan}
\label{app:plan}

\paragraph{Contents.}
The VLM sees a grid of nine frames, each labelled with its frame index and with
the tracks' boxes and names, together with the question and the numbered prior
statements. It returns the objects in build order (supports first, then what rests
on them, then what is held or flies), each bound to a track or to none, with a
shape kind, proportions $L\!:\!W\!:\!H$ as ratios, the view in which the camera
sees it, and its motion as a sequence of states, each with a frame range, a
relation and its target, a law, and a direction (left, toward the camera, turning,
and so on). The plan also binds every prior statement and the question, names the
environment, and says whether the camera moves. A second call reviews the plan. It
may add an object or sharpen a description, but may not remove, rename, rebind,
or reorder objects.

\paragraph{Re-asks and repairs.}
The plan is checked in code against a fixed set of rules. A plan that breaks a
rule is returned to the model once with the reason. If the reply still breaks it,
the rule is applied deterministically and the repair is recorded. Among the rules:
\begin{itemize}
  \item A statement binds only to a tracked object.
  \item A statement about a class (``the
tennis balls are 6.7\,cm'') binds to every tracked instance of the class.
  \item A size
question binds to the axis the view shows, so the ``width'' of a body seen
side-on is its length $L$.
\end{itemize}

\paragraph{Statement accounting.}
Every prior statement has exactly one account. It is a binding to a parameter or derived
quantity, an image measurement, or an explicit declaration that nothing observable
carries it, with the reason. No statement is dropped silently.

\paragraph{Depth groups, class sizes, and image measurements.}
A depth group names bodies at about the same distance from the camera, at a given
frame or, for bodies that do not move relative to one another, over their whole
common presence. Each group becomes same-depth ties between its members, and code
rejects a tie that contradicts the prior's distance statements. A class size is a
commonsense size of a tracked body with its spread, which becomes a soft vote. A
vote on the quantity the question asks for is dropped, and a prior statement on
the same quantity replaces it. An image measurement binds a statement about a
feature no track carries to two end points that the model marks on one grid frame,
on the plane the feature lies on, and is back-projected through the current camera
onto that plane. For a repeated feature, the model marks the two farthest lines it
can still count without doubt and gives the number $n$ of periods between them.
The measured length is $n$ times the stated spacing, so the pixel error of the end
points is divided by $n$.

\paragraph{Shapes.}
For each body, one model call writes its OpenSCAD module from crops of the body.
The module consists of CSG primitives in the body's own frame, normalized to the unit box, whose metric
extents $(L,W,H)$, or a diameter $D$, are scene parameters.
The module must use only whitelisted primitives (cube, sphere, cylinder, their
transforms, and Boolean operations), compile, form one connected solid, and have
extents whose order agrees with the planned proportions. A failure is returned
with its measured reason (the compiler message, the number of components, the
measured extents), at most three times. A body may be a part of another body, such
as the deck of a pier. It is then placed at a fixed offset inside its whole and
solved together with it.

\paragraph{Shape check.}
Each body's module is rendered nearly orthographically at its own ratios from yaw
angles every $15^\circ$ and from elevations that follow the planned view. On up to
five evenly spread frames on which the mask is neither cut by the border nor
shared with another track, each direction's silhouette is aligned to the mask by
its centroid, scale, and principal axis, and the best direction is kept per frame.
The check reports the boundary distance after alignment (median over the frames),
that distance relative to the mask's equivalent radius, and the rendered and
observed extents along the mask's principal axes. One VLM call per round sees, for
every body, crops with the observed and the aligned outlines and these numbers,
and returns for each body either ok or a fix. A fix re-authors the module or
changes the proportions on the axes the view shows. A sphere is not judged, nor is
a body without a track. A fix is kept only when the body's boundary distance falls
by more than the larger of $\sigma_o$ and $\sqrt{2}$\,px, and the plan and module are
otherwise restored exactly. A body stops when it is judged ok, when no untried fix
remains, after three rounds, or when the question's time budget of 120\,s is
spent.

\paragraph{Structure and question.}
Parameters have values only where the prior or a convention gives them. A prior
statement on a size or on $g$ fixes that parameter. A statement on a derived
quantity becomes a held row. The question is parsed in code, without a model,
into the quantity asked, a time or window, an aggregate (average, start, end,
maximum, minimum, total), a horizontal or vertical component, and the unit.

\subsection{Objects, Motion Laws, and Camera}
\label{app:laws}

\paragraph{Relations.}
Each segment of an object's presence has a relation to an earlier object.
The relation is \emph{free}, \emph{on}, \emph{in}, \emph{held by}, or \emph{hangs from}. The
relation fixes the frame in which the segment's law is written. For \emph{on}, the
object's height comes from the top face of its support instead of being
estimated. A body that moves by a law \emph{on} another body stays within the
support's footprint. Its contact point lies inside the footprint at both ends of
the segment. When the next segment is a ballistic flight entered with a
continuous velocity, the contact point at the switch lies on the edge its path
crosses, which also measures the support's extent. These terms are tested on the
scene fitted without them. Those that agree with the plan up to three frames of
boundary freedom are held in one more joint pass. A term that fails contradicts
the plan, so the support is not distorted and the departure is recorded for the
path check.

\paragraph{Laws.}
With $\tau=t-t_0$ the time since the segment's start, the laws are
\begin{align*}
  \text{static:}\quad & \mathbf{p},\\
  \text{constant velocity:}\quad & \mathbf{p}_0+\mathbf{v}\tau,\\
  \text{constant acceleration:}\quad & \mathbf{p}_0+\mathbf{v}_0\tau+\tfrac{1}{2}\mathbf{a}\tau^2,\\
  \text{ballistic:}\quad & \mathbf{p}_0+\mathbf{v}_0\tau-\tfrac{1}{2}g\,\tau^2\hat{\mathbf{z}},\\
  \text{rolling:}\quad & \mathbf{p}_0+\hat{\mathbf{v}}_0\big(|\mathbf{v}_0|\tau'-\tfrac{1}{2}d\,\tau'^2\big),\\
  & \tau'=\min(\tau,\,|\mathbf{v}_0|/d),
\end{align*}
a \emph{pendulum} given by its pivot, length, initial angle and angular velocity,
swing plane, and damping, integrated with the nonlinear equation of motion.
A \emph{circular} motion is given by its center, radius, initial angle, and angular
velocity. A rolling body moves along its support with deceleration $d$, comes to
rest when its speed reaches zero, and spins about its center with radius $H/2$.
The \emph{driven} law follows a table of per-frame poses. For \emph{on}, the law is
written in the top face of the support and its vertical component is dropped. The
heading is a fixed yaw, the direction of the velocity, a fixed 3D rotation, or a
table. A ballistic segment (Eq.~\eqref{eq:ballistic}) has six parameters,
$\mathbf{p}_0$ and $\mathbf{v}_0$, however many frames it spans. The driven law can
follow any motion but gives no speed or acceleration. The scene parameter $g$ is
read by the ballistic and pendulum laws. Environments set $g$ by convention, with $9.81\,\mathrm{m/s^2}$ on Earth,
$1.62\,\mathrm{m/s^2}$ on the Moon, and zero in free fall.

\paragraph{People.}
The person template of Figure~\ref{fig:person} takes every length as a fixed
fraction of the standing height $H$ from the segment table of
\citet{drillis1964body} as tabulated by \citet{winter2009biomechanics}. The head
is $0.130\,H$, the shoulders stand at $0.818\,H$ and are $0.259\,H$ wide, the
hips are $0.191\,H$ wide, and the upper arm, forearm, thigh, and shank are
$0.186\,H$, $0.146\,H$, $0.245\,H$, and $0.246\,H$ long. The head is a sphere,
the torso a hull of two slabs, and each limb a box with rounded ends, fifteen
solids in all. The mesh is normalised to the standing height in every pose, so
$H$ is the standing height also for a seated or crouching person. A body is a
candidate when its name or class contains a human word and no excluding word,
such as robot, mannequin, statue, or animal. One model call per question then
confirms each candidate and names its pose per motion state. The pose kept is the
one of the state that holds the asked time, else the majority over the frames.
The proportion gate and the shape check never edit a person.

\paragraph{Camera.}
The camera is a static pinhole with focal length $f$, the principal point at the
image center by convention, and a pose given by its height, pitch, and roll above
the first plane, usually the ground. The world's $z$ axis points up and the first plane is $z=0$. The camera's horizontal
position and yaw are gauges fixed by convention. The program renders every object
on every frame with occlusion by nearer objects.

\subsection{Fitting}
\label{app:fit}

\paragraph{Silhouette rows.}
For object $o$ on frame $f$, a fixed set of points on the shape's surface is
projected through the program. Precision terms $p_{ofj}$, $j=1,\dots,G$, are the
root-mean-square distances of groups of projected points to the region the object
may occupy. This region is its mask, extended by pixels a nearer modelled object covers, by
tracks the program does not model, and by the outside of the image where the mask
touches the border. Recall terms $r_{ofk}$, $k=1,\dots,K$, are taken at sampled
points on the observed silhouette's edge. Each is the signed extent of the
projected shape along the edge's outward normal. The discrepancy of
Eq.~\eqref{eq:objective} is
\begin{equation}
  d_{of}=K_{\mathrm{e}}\Big(\frac{1}{G}\sum_{j=1}^{G}p_{ofj}^2+\frac{1}{K}\sum_{k=1}^{K}r_{ofk}^2\Big),
\end{equation}
so each frame counts as $2K_{\mathrm{e}}$ samples. The objective is minimized with
a trust-region least-squares solver. The pull toward initial values is measured in
natural units. The occlusion context is rendered from the current program and held
fixed within a solve.

\paragraph{Held rows.}
The held rows are the prior's statements on derived quantities (speed,
acceleration, distance to camera, distance between objects, height above a
support), image measurements with a stated value, continuity of position across
consecutive segments (and of velocity where both laws imply it), same-depth ties,
and contacts, where an object leaving or reaching a holder unobserved is at the
holder's depth at that moment.

\paragraph{Projection.}
After every change of the parameters, the parameters that a group of rows reads
are projected onto the set where all rows of the group hold, by minimum-norm
Newton steps that stop at relative residual $\epsilon_0$. Objects whose rows read
one another's poses form one group and are projected together, in lexicographic
tiers, first the object being fitted, then the other objects of the group, and the
camera only for what no object can satisfy. A final step projects every set of
violated rows once more from the final state over all the variables they read, and
keeps the result where every row then holds.

\paragraph{Penalty rows.}
Where the projection cannot bring a row to its target at a trial point, the
solver must not move away from the row for a better silhouette at no cost. Every
solve therefore carries, for each held row of the projected groups,
\begin{equation}
  \label{eq:penalty}
  \pi_h=\operatorname{sign}(\epsilon_h)\,\frac{\max\!\big(|\epsilon_h|-\epsilon_0,\,0\big)}{\tilde\tau_h},
\end{equation}
added as $\sum_h\pi_h^2$ to the objective, with $\epsilon_h$ and $\tilde\tau_h$ the
row's residual and tolerance relative to its scale. A row the projection holds
costs nothing. Any other row costs one noise unit per tolerance, inside the
tolerance as well as beyond it.

\paragraph{Tolerances.}
Table~\ref{tab:tolerance} lists how each row's tolerance is read from the problem.
Two statements on the same quantity of the same object at the same time whose
values differ by more than their combined written precision are both set aside as
contradictory, with the reason. Velocity continuity is evaluated with second-order
one-sided differences, which are exact for every law of at most second order in
time. A first-order difference would read an accelerating segment's start as a
spurious velocity jump.

\begin{table}[t]
  \centering
  \caption{Tolerance $\tau_h$ of a held row. $Z_o(t_h)$ is the object's depth at
  the row's time and $f$ the focal length.}
  \label{tab:tolerance}
  \small
  \begin{tabular*}{\columnwidth}{@{\extracolsep{\fill}}p{0.34\linewidth}p{0.58\linewidth}@{}}
    \toprule
    Stated value & half a unit of the last written digit, in SI (``57.2\,mm'' gives 0.05\,mm)\\
    Position continuity & $\sigma_o Z_o(t_h)/f$ across the line of sight\\
    Velocity continuity & the same, divided by the shorter segment's duration\\
    Same-depth tie, contact & the two objects' tolerances in quadrature\\
    Image measurement & the larger of the written precision and the pixel $\sigma$ of the back-projected length\\
    \bottomrule
  \end{tabular*}
\end{table}

\paragraph{Order of the fit.}
The focal length starts from the monocular metric depth model Depth
Pro~\cite{depthpro2025}, the median of its per-frame estimates on 12 evenly spaced
frames of the video, and is a free variable of the fit from there. The focal
length is held at its start while the objects are fitted one by one, until the joint pass below frees it.
The first tracked body is fitted together with the camera's pitch, roll, and
height from each pitch start. A start is invalid if it puts the camera below a
plane on which a tracked object rests, or if the prior's statements cannot hold at
that camera. Each valid start is refined and scored on the whole scene. Every
start within one noise unit per frame of the best, and at least three, is carried
through the complete fit, and the lowest final objective is kept. The other bodies
follow in build order with the camera and the earlier bodies frozen, each from
discrete heading starts. A joint pass then refines all of them and frees the
focal length. The fitted value is kept only if the silhouettes and the prior
determine it (relative $\sigma \le 0.25$, Table~\ref{tab:constants}) and it does not
reach the bound of the physical field of view. Otherwise the focal length is reset
to its start, the joint pass is repeated with it fixed, and it is reported as
undetermined. Keeping the fitted value does not vouch for it. Whether the focal
length is reported as determined is decided, as for every parameter, by the
identification after the fit (Appendix~\ref{app:ident}). A body's size starts as one
scale times the model's ratios. After the joint pass, each extent that the
observations determine on its own is freed, and the others remain ratio times
scale and are reported as undetermined. Segment boundaries may move by up to three
frames.

\paragraph{Gates.}
Each fitted object passes or fails four gates. (1) The program evaluates. (2) Its
held rows hold to their tolerances. (3) Its relations hold. (4) Its pose agrees
with the masks. For the pose gate, the 2D similarity (translation, scale,
rotation) that aligns the observed mask with the rendered one on each frame is
summarized per segment as a constant part, a drift over the segment, and a scatter
about that drift, each in boundary pixels. A constant offset points to a wrong
size or placement. A drift or scatter points to a wrong law or segment boundary.
The gate is size-aware for the constant part only. The silhouette of a coarse
solid may lie $0.1\,R$ from its mask, with $R$ the equivalent radius of the
observed silhouette on the segment, so the constant part passes up to
$\max(\sigma_o,\sqrt{2}\,\mathrm{px},0.1\,R)$. The drift and the scatter get no such
allowance, because a few pixels of scatter on a large object are a large error in
its speed at an instant. Their bound is
$\max(\sigma_o,\sqrt{2}\,\mathrm{px},\min(r_s,0.1\,R))$, where $r_s$ is the segment's
shape residual, the solid's own mismatch with the mask. The shape residual left
after alignment is reported but does not gate.

\subsection{Identifiability and the Read-out}
\label{app:ident}

\begin{table}[t]
  \caption{\textbf{Determined and undetermined parameters} for the question of Figure~\ref{fig:pipeline}. Every parameter has a value, but only filled ones are vouched for, with their $\sigma$. The camera's roll is determined. Its focal length, height, and pitch are left open, and with them the black and purple balls' positions along the table, the black ball's motion, and the asked distance.}
  \label{tab:ident}
  \begin{tabular*}{\columnwidth}{@{\extracolsep{\fill}}llcl@{}}
    \toprule
    Object & Parameter & & Fitted value \\
    \midrule
    % written by figures/src/identifiability.py from fit.json; do not edit by hand
camera & f & \textcolor{qpGrey}{$\circ$} & $1681$\,px \\
 & height & \textcolor{qpGrey}{$\circ$} & $1.54$\,m \\
 & pitch & \textcolor{qpGrey}{$\circ$} & $62.3^\circ$ \\
 & roll & \textcolor{qpGreen}{$\bullet$} & $-1.8 \pm 4.0^\circ$ \\
\midrule
black ball & x & \textcolor{qpGrey}{$\circ$} & $0.72$\,m \\
 & $|v_0|$ & \textcolor{qpGrey}{$\circ$} & $0.055$\,m/s \\
 & deceleration & \textcolor{qpGrey}{$\circ$} & $0.050$\,m/s$^2$ \\
\midrule
purple ball & x & \textcolor{qpGrey}{$\circ$} & $0.86$\,m \\
\midrule
answer & distance & \textcolor{qpGrey}{$\circ$} & $15.32 \pm 2.65$\,cm \\
\bottomrule
\multicolumn{4}{@{}l@{}}{\rule{0pt}{2.6ex}\textcolor{qpGreen}{$\bullet$}\ determined\qquad \textcolor{qpGrey}{$\circ$}\ undetermined} \\

  \end{tabular*}
\end{table}

\paragraph{Uncertainty estimates.}
In Eq.~\eqref{eq:covariance}, class-vote rows are multiplied by $s$ so that a vote
enters with its stated $\sigma_c$. A parameter's $\sigma$ is the largest of four
estimates. They are (1) the classical covariance of Eq.~\eqref{eq:covariance}, (2) a
frame-clustered sandwich covariance, robust to correlated residuals within a
frame, (3) a profile along the weakest directions, with the other parameters
re-optimized, and (4) the spread of the parameter over the equally good starts of
the fit.

\paragraph{Undetermined parameters.}
A parameter is undetermined if $\sigma$ exceeds a quarter of its natural scale
(its own magnitude for a size or a focal length, the object's distance for a
position, and for a velocity the larger of its magnitude and the object's length
over the segment duration), if it loads a direction of zero curvature, if it has
no effect on any frame or row, or if it ends at a bound.

\paragraph{Scale test.}
Every length, speed, and acceleration that no metric statement anchors is
multiplied by $1.01$ about the world origin, while the anchors (the prior's
statements and fixed sizes, $g$, image measurements, and class votes) are held.
If neither the silhouette objective nor any statement changes, the scale is
undetermined. Otherwise each anchor is tested alone.

\paragraph{The read-out's $\sigma$.}
The covariance is propagated through the quantity's gradient. The written
precision of each statement it rests on is added, and the result is enlarged by
the misfit of the frames it is read from and by pose residuals above the gate
threshold, converted to the asked quantity (a drift of $d$\,px over a segment of
duration $T$ at depth $Z$ is a speed error of $dZ/(fT)$). An image measurement
asked for is read per period.

\paragraph{Empty read-outs.}
The read-out is empty, with a named reason, in each of these cases:
\begin{itemize}
  \item A parameter it reads, or a component the fit does not vary, is open for a
structural reason along which its value changes. Such reasons are a direction of
zero curvature, an unbounded profile, the global scale, a held or bounded
variable, a table cell, or a parameter without observation.
  \item Its value differs, beyond its $\sigma$, over the equally good states of
the fit.
  \item Its own uncertainty from the observations, without the prior's written
precision, exceeds 0.25 of its natural scale. In the reported run, 40 of the 111
empty read-outs are empty for this reason alone, such as a speed of 4.0\,m/s with
$\sigma = 1.02$\,m/s, 0.254 of its scale.
  \item It is a speed or
acceleration read where the motion is a pose table.
  \item A held row is violated in
the final fit, since all objects share the camera through their silhouettes.
  \item The scale rests only on a class vote that the fit pulls beyond $3\sigma_c$.
  \item The question is bound to a quantity the program has no expression for.
\end{itemize}

\paragraph{Extrapolated read-outs.}
When the asked time lies outside the frames in which a tracked body is observed,
the read-out evaluates the law of the segment that holds the time, else of the
body's last or first segment, beyond the segment's end or before its start. A
pose table is not a law and gives no value. The law is continued only while it
stays valid. A free body that comes down on a plane or on a support's top face
has left its law, since no bounce is modelled, and so has a body whose contact
point leaves its finite support. The read-out is then empty and records the time
and the surface. The $\sigma$ adds the read segment's pose residual as a velocity
error times the continued duration, and the determinacy rule is unchanged.

\subsection{Answer Step}
\label{app:answer}

The final-answer call is one request. It receives 16 frames sampled uniformly
over the video, resized to a long side of at most 768\,px, each preceded by its
frame index and time. It also receives up to five overlays of the rendered program
on the video, each labelled with every object's IoU against its mask. The text
gives the question, the asked unit, the prior, and the reconstruction. The
reconstruction lists each object with its shape, its sizes, and, for every
segment, its law, relation, and fitted values, each with the source that gave it
(the prior, the fit, a convention, or the model's guess). It also lists $g$, the
camera, what holds the metric scale, each prior statement beside the value the
reconstruction has for it, and the reconstruction's value for the question. The
prompt says that the reconstruction can be right or wrong. It carries no
determined flag, no $\sigma$, and no verdict of the read-out. The call judges
whether the reconstruction is credible for the question and returns one number
in the asked unit with a rationale and the inputs it relied on. Each prior
statement is listed with the reconstruction's value converted to the statement's
own unit and, for a counted feature, per period.

\section{Experimental Details}
\label{app:exp}

\paragraph{Metric.}
For a prediction $\hat{y}$ and ground truth $y$, MRA is the fraction of thresholds
$\theta\in\{0.1,0.2,\dots,0.9,0.95\}$ for which the relative error
$|\hat{y}-y|/|y|$ is below $1-\theta$. A missing answer scores zero.

\paragraph{\model.}
Each question is run once, on its own, with the constants of
Table~\ref{tab:constants}, on the setup of Section~\ref{sec:exp-implementation}.
Table~\ref{tab:timing} gives the time of each step per question. The plan, the
authoring, the shape check, and the final answer make 7 model calls per question
(median, 4 to 18). Perception adds the calls of the object list, the choice of detection boxes,
and the visibility labels of the presence check.

\paragraph{Shape check.}
Every reported program went through the shape check of
Section~\ref{sec:method-program}. It measured all 495 bodies of the 158 programs,
and the VLM judged 272 of them. It proposed 62 changes and kept 5, one body on
each of 5 questions. The other 57 did not lower the boundary distance by more than
the mask noise and were reverted.

\paragraph{Scale anchors.}
Table~\ref{tab:anchors} groups the questions by the scale test of the fit. It reads the field \texttt{held\_by} of the scale entry of the fit's
identifiability report, which is \emph{prior} when only prior statements, prior
sizes, and image measurements of prior features fix the scale (127 questions,
7 of them through an image measurement), and \emph{prior and convention} when $g$
taken as the environment's convention fixes it as well (31). The scale is
determined on every question.

\begin{table}[t]
  \centering
  \caption{\textbf{Time per question} over every question with a run, in the
  order of the steps. The time of the final answer includes the queue of the
  shared model endpoint. The focal length start is computed once per video, so it
  costs nothing on most questions.}
  \label{tab:timing}
  \small
  \begin{tabular*}{\columnwidth}{@{\extracolsep{\fill}}lrrr@{}}
    \toprule
    Step & Median & Max & Total \\
    \midrule
    Object list    & 22\,s    & 40\,s   & 1.0\,h  \\
    Labelled frame grid & 6\,s & 34\,s  & 0.3\,h  \\
    Detection, tracking, presence & 159\,s & 17\,min & 8.3\,h \\
    Focal length start & 0\,s & 54\,s  & 0.1\,h  \\
    \midrule
    Masks          & 38\,s    & 100\,s  & 1.7\,h  \\
    Plan, shapes, structure & 99\,s    & 330\,s  & 4.9\,h  \\
    Fit            & 21\,min  & 5.3\,h  & 99.0\,h \\
    Final answer   & 17\,s  & 103\,s   & 1.0\,h  \\
    \bottomrule
  \end{tabular*}
\end{table}

\paragraph{Code as Worlds.}
Both sizes are run with the authors' released evaluator on 16 frames, at
temperature 0.01 with a fixed seed and without thinking, on one NVIDIA L40S. Only
the GPU memory fraction is raised from the released setting, which is sized for
80\,GB GPUs.

\paragraph{PhysMind.}
We run PhysMind's own stages, namely SAM~3~\cite{sam3_2026} tracks from the VLM's concept prompts naming
the question objects, MoGe-2~\cite{moge2}, Metric Video Depth Anything~\cite{videodepthanything2025},
SAM~3D Objects~\cite{sam3dobjects}, FoundationPose~\cite{foundationpose2024},
and GeoCalib~\cite{geocalib2024}.
The asked quantity is measured on the posed metric meshes without the prior, as
in PhysMind. Of the 158 questions, 90 are measured this way. For the other 68
(no posed track of the question object in 46, no detection in 15, no poses in 7)
the answer is the VLM's direct answer on the whole video, as PhysMind's own
pipeline does. The VLM is Gemini~3 Flash (preview)~\cite{gemini3flash}. Tracks that SAM~3 returns
without a tag are matched to the question objects by the same VLM. The PhysMind panels of Figure~\ref{fig:qualitative} are
its posed meshes from this re-run of its released code, drawn onto the video frame.

\paragraph{Thinking with Spatial Code.}
We run the released code (commit cf7fffc) with the released Spatial Encoder
weights on 32 frames spread over the video. Gemini~3 Flash boxes each object on
the earliest of 8 shown frames in which it is visible, and the encoder turns the
boxes into the spatial code. The code holds one 3D box per object and its per-frame box centres
with timestamps, in metres in the first camera frame. The reasoner,
Qwen3-4B-Instruct-2507~\cite{qwen3} at temperature 0 with at most 8192 output tokens, receives
the spatial code, the prior, the given distances, and the question. Everything
runs on one NVIDIA L40S. The Thinking with Spatial Code panels of
Figure~\ref{fig:qualitative} are its 3D boxes from this run, each projected with
the encoder's own camera of the sampled frame nearest to the frame shown.

\paragraph{Direct inference.}
Gemini-4-Carbon~\cite{gemini4carbon} and Qwen3.5-27B~\cite{qwen35} are queried without tools. Each receives the video
file itself, with no frame sampling on our side, and one prompt with the prior,
the given distances for 3D questions, and the question, asking for the number
alone. Both were run on 2026-11-13 through OpenRouter on the model owner's own
hosts. Gemini-4-Carbon runs as \texttt{google/\allowbreak gemini-4-carbon} with its default
reasoning setting. On 11 of the 158 questions it returned no answer, and these
score zero. Qwen3.5-27B runs as \texttt{qwen/\allowbreak qwen3.5-27b} with
thinking enabled. Its 138 questions on 20 videos ran on Alibaba's hosts with no
fallback to another provider. These hosts rejected the video input of the other
four videos with an invalid-parameter error. The 20 questions of those videos were
run again on 2026-11-14 through OpenRouter without pinning the provider. On one
question the model returned an empty answer, which scores zero. On the 157
questions it answered, its MRA on D2 is 0.503 (36 questions). The other categories
are as in Table~\ref{tab:quantitative}.

\begin{figure}[t]
  \centering
  \includegraphics[width=\linewidth]{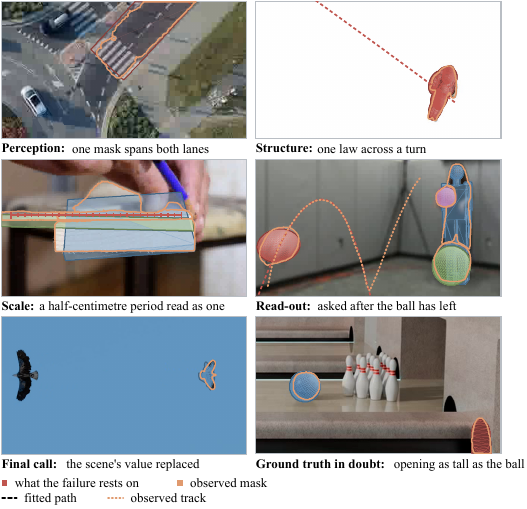}
  \caption{\textbf{Failure cases.} The 24 questions whose final answer scores below 0.5 fall into six groups, each question assigned to the earliest stage at which it goes wrong. The groups are perception (2), structure (8), scale (2), read-out (6), final call (2), and ground truth in doubt (4). One question is shown for each group. The examples are a lane mask that spans both lanes, an astronaut who turns under one constant acceleration (dashed), a ruler period of half a centimetre read as one (red), a ball asked about long after its track (dotted) has left the frame, a correct speed of the small bird replaced by a pixel estimate, and a ball return opening about as tall as the 0.21\,m ball, released as 10\,m. The final program is drawn over each frame.}
  \Description{Six video frames in a three-by-two grid, each with the reconstructed scene drawn over it and a one-line label naming the stage (perception, structure, scale, read-out, final call, and ground truth in doubt).}
  \label{fig:failures}
\end{figure}

\subsection{In-the-Wild Videos}
\label{app:wild}

The 27 videos fall into three groups. They are 11 ball videos (robot soccer and
basketball, with the regulation ball diameter as the prior), 6 videos of
quadruped robots (with the manufacturer's published body length), and 10 lab
recordings of objects that are pushed or tossed (with the length of a standard
ceiling light panel). Every video has a static camera and an object of standard
size. Each video has two questions,
written from the frames before any run, and no ground truth.

The run used the setup of Section~\ref{sec:exp-implementation}. Of the 54
questions, 45 were answered. On all 45, the scale rests on the stated size alone
on 28 and on the stated size together with $g$ on 17. The answers are not scored.
Figure~\ref{fig:wild} shows five of these questions.

\section{Failure Cases}
\label{app:failures}

Because every intermediate quantity is explicit, each error can be traced to the
stage that caused it (Figure~\ref{fig:failures}). We assign every question whose
final answer scores below 0.5 to the earliest stage at which it goes wrong, from
the masks to the final call, by inspecting the frames, the tracks, the rendered
program, the fit, and the final call's rationale. Of the 158 final answers, 24
score below 0.5. They fall into six groups. In four of them, perception,
structure, scale, and read-out, the reconstruction goes wrong. In the fifth, the
final call, the reconstruction is right and the final call is not. In the sixth,
the reconstruction agrees with the video but not with the released value.

\emph{Perception}, 2 questions. The mask or track belonged to a different or
larger region than the named object, such as a whole road approach or a nearby
walker, and the fit then measured that region faithfully.

\emph{Structure}, 8. One law was fitted over a motion that changes, or a segment
boundary, a support, or a pivot was placed where the video does not put it.

\emph{Scale}, 2. The anchor reading misjudged what the reference marks measure.
A ruler's half-centimetre period was read as one centimetre, for example, so every
length inherited the same factor.

\emph{Read-out}, 6. The program matched what it observed, but the asked quantity
lay outside it. It lay after the body left the frame, along the wrong axis, in a
driven pose table, or in a relative depth that nothing observed.

On 3 of the 24, two in structure and one in ground truth in doubt, the final
call kept a determined scene value to within 1\%
(Section~\ref{sec:exp-ablations}).

\emph{Final call}, 2. The reconstruction offered a value close to the truth, but
the final call replaced it with a rough pixel measurement of its own. This is a
limitation of the VLM that reads the scene, not of the reconstruction.

\emph{Ground truth in doubt}, 4. The released value contradicts what the frames
show directly, such as object sizes, stated distances, or visible motion. These
questions are scored against the released value like every other question, and
Appendix~\ref{app:gt-doubt} describes them.

\subsection{Ground Truth in Doubt}
\label{app:gt-doubt}

On the four questions of this group, the reconstruction is consistent with the
video but not with the released value. In the example of
Figure~\ref{fig:failures}, the ball return opening is about as tall as the
0.21\,m ball beside it, while the released height is 10\,m.

\section{Limitations}
\label{app:limitations}

\paragraph{A static pinhole camera.}
The program has one static pinhole camera with the principal point at the image
center and no lens distortion. A video that pans, zooms, or follows a body breaks
this assumption, and the fit then explains the image motion the camera causes as
motion of the objects or as misfit.

\paragraph{Coarse solid shapes.}
Every body is a connected solid of a few primitives with metric extents along its
own axes. A body that deforms or articulates is matched only on average, and a
part that the coarse shape does not separate cannot be measured.

\paragraph{The scale needs an anchor.}
Masks fix a scene only up to a common scale. Without a prior statement, $g$ in a
free-flight or swinging segment, or a class size, every metric read-out is
undetermined. A class size is only as good as a commonsense size, and a wrong
environment gives a wrong $g$.

\paragraph{The structure depends on the VLM\@.}
The objects, their shapes, supports, segments, and laws, and what each prior
statement refers to are written by the VLM. Code checks the plan against rules,
and the fit exposes errors that show in the measurements, but a wrong structure
that fits the masks as well as the right one is not detected.

\paragraph{Fitting time.}
The fit is an optimization per question, not a forward pass. The method is
therefore suited to offline measurement rather than to interactive use.

\paragraph{Empty read-outs.}
The read-out has no value when the evidence does not determine the asked
quantity. This is by design, so the read-out covers only part of the questions.
The final answer is always given.

\end{document}